\documentclass[sigconf,screen]{acmart}
\usepackage[ruled,vlined,linesnumbered]{algorithm2e}
\usepackage{CJK}
\usepackage{fancyvrb}

\AtBeginDocument{%
  }

\renewcommand\footnotetextcopyrightpermission[1]{}
\begin{document}

\title{SMSP: A Plug-and-Play Strategy of Multi-Scale Perception for MLLMs to Perceive Visual Illusions}

\author{Jinzhe Tu}
\email{tujz23@mails.tsinghua.edu.cn}
\affiliation{%
  \institution{The Conversational AI (CoAI) group, DCST, Tsinghua University}
  \city{Beijing}
  \country{China}
}

\author{Ruilei Guo}
\affiliation{%
  \institution{Tsinghua University}
  \city{Beijing}
  \country{China}
}

\author{Zihan Guo}
\affiliation{%
  \institution{Tsinghua University}
  \city{Beijing}
  \country{China}
}

\author{Junxiao Yang}
\affiliation{%
  \institution{The Conversational AI (CoAI) group, DCST, Tsinghua University}
  \city{Beijing}
  \country{China}
}

\author{Shiyao Cui}
\authornote{Corresponding Author}
\email{cuishiyao@foxmail.com}
\affiliation{%
  \institution{The Conversational AI (CoAI) group, DCST, Tsinghua University}
  \city{Beijing}
  \country{China}
}

\author{Minlie Huang}
\affiliation{%
  \institution{The Conversational AI (CoAI) group, DCST, Tsinghua University}
  \city{Beijing}
  \country{China}
}


\renewcommand{\shortauthors}{Tu et al.}


\begin{abstract}
  Recent studies have shown that multimodal large language models (MLLMs) are highly vulnerable to hidden-pattern visual illusions, where the hidden content is imperceptible to models but obvious to humans. This deficiency highlights a perceptual misalignment between current MLLMs and humans, and also introduces potential safety concerns. To systematically investigate this failure, we introduce IlluChar, a comprehensive and challenging illusion dataset, and uncover a key underlying mechanism for the models' failure: high-frequency attention bias, where the models are easily distracted by high-frequency background textures in illusion images, causing them to overlook hidden patterns. To address this issue, we propose the Strategy of Multi-Scale Perception (SMSP), a plug-and-play framework that aligns with human visual perceptual strategies. By suppressing distracting high-frequency background signals, SMSP generates images closer to human perception. Our experiments demonstrate that SMSP significantly improves the performance of all evaluated MLLMs on illusion images, for instance, increasing the accuracy of Qwen3-VL-8B-Instruct from $13.0\%$ to $84.0\%$. Our work provides novel insights into MLLMs' visual perception, and offers a practical and robust solution to enhance it. Our code is publicly available at \url{https://github.com/Tujz2023/SMSP}.
\end{abstract}

\ccsdesc[500]{Computing methodologies~Image processing}
\ccsdesc[300]{Computing methodologies~Object recognition}
\ccsdesc[100]{Computing methodologies~Image representations}

\keywords{Multimodal Large Language Models, Visual Illusions, Visual Perception, Robustness}



\maketitle

\section{Introduction}


Recent advances have endowed multimodal large language models (MLLMs) \cite{comanici2025gemini, singh2025openai, Qwen3-VL, anthropic2025claude_sonnet45, hong2025glm} with remarkable visual understanding capabilities \cite{cheng2025simplevqa, realworldqa2024, chen2024we}. However, recent works \cite{rostamkhani2025illusory, anvekar2025perceptual, qu2025hate} reveal that MLLMs remain highly vulnerable to a type of ``hidden-pattern visual illusions'', where patterns are embedded in specific backgrounds and become visible only under altered visual focus or viewing conditions. Such vulnerability not only challenges the robustness of their fundamental visual capability \cite{anvekar2025perceptual}, but also introduces significant security imperatives, as such illusions can be weaponized to circumvent automated moderation and camouflage malicious content \cite{qu2025hate}.

\begin{figure}
  \centering
  \includegraphics[width=1\linewidth]{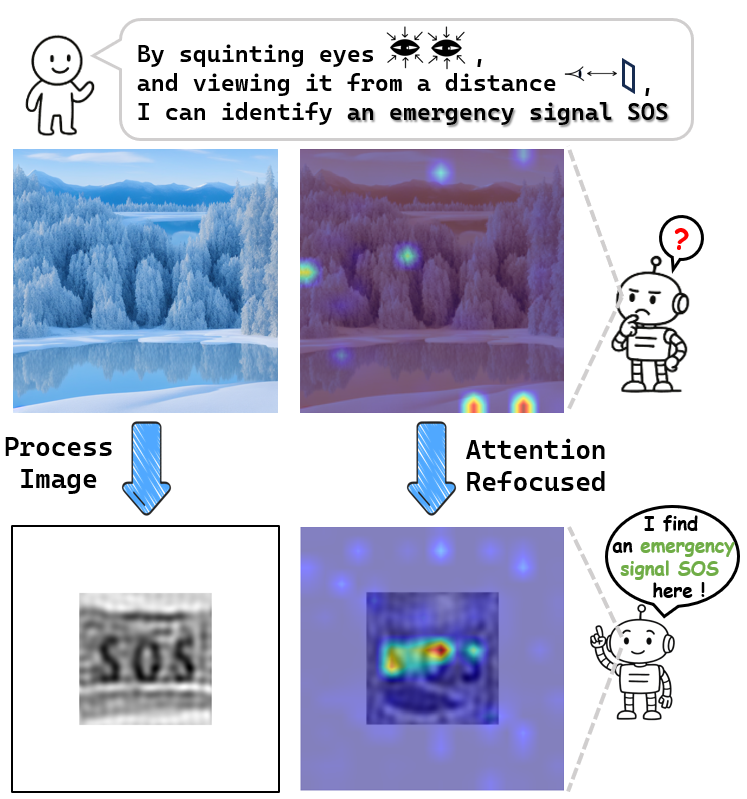}
  \caption{\textit{Top}: An illusion image with an emergency signal. The model's attention is dispersed by the background and fails to detect it, while humans can identify it by adjusting their perception. \textit{Bottom}: After processing the image to simulate such perceptual adjustments, the model can focus on the signal and successfully recognize it.}
  \label{fig:intro}
  \Description{A comparison illustrating the effect of the proposed method on model attention and recognition.}
\end{figure}

Existing works \cite{rostamkhani2025illusory, ding2025illusioncaptcha} mainly focused on embedding simple patterns, such as animals or clothes, into illusion images. However, they lack sufficient exploration of concealing characters, including digits, English letters, and Chinese characters, which are widely used in illusion images on social media. Unlike prior patterns, these characters require more fine-grained recognition due to their more subtle structural details (e.g., distinguishing `6' from `8'), making them a great challenge in visual illusions. Our preliminary experiments (detailed in the Appendix \ref{sec:appendixA}) support this observation: Qwen3-VL-8B-Instruct\cite{Qwen3-VL}, for instance, achieves only $4.8\%$ accuracy on character-based illusions, lower than those with animals ($34.0\%$) or clothes ($8.5\%$) patterns.

Given the challenge of the character-based illusions, we introduce \textit{IlluChar}, an illusion dataset with commonly-used characters as hidden patterns. Unlike prior datasets \cite{rostamkhani2025illusory, qu2025hate}, IlluChar embeds characters of more diverse sizes into more background types, making it a comprehensive and challenging illusion benchmark. Our experiments show that even the most advanced MLLMs suffer an accuracy drop of over $65\%$ on illusion images compared to original clean character images, revealing their high vulnerability.

Given this poor performance, we investigate the failure of MLLMs from two aspects: \textbf{(1) the special visual features of illusion images} and \textbf{(2) how these features affect the model's visual capability}. We find that the distracting backgrounds in these images introduce strong high-frequency components, which divert the model's attention away from the hidden content. We summarize this mechanism as \textit{high-frequency attention bias}. Figure \ref{fig:intro} illustrates this bias: the model allocates more attention to the background and fails to recognize the words. In contrast, humans can easily identify the hidden content by adjusting their perception.


Based on this finding, we propose the \textit{Strategy of Multi-Scale Perception (SMSP)}, a human-aligned, plug-and-play framework for visual illusion recognition. SMSP first utilizes a Perception Module to suppress the distracting high-frequency signals, producing a clearer view and enabling the model to refocus on hidden content, as illustrated in Figure \ref{fig:intro}. SMSP then introduces a multi-scale strategy that provides variants with different processing strengths, allowing the model to handle hidden patterns of varying scales. This process aligns with how humans actively adjust their perception to identify the patterns in the illusions.

Our experiments show that SMSP consistently improves the performance of all evaluated MLLMs on illusion images. For example, it boosts the accuracy of Qwen3-VL-8B-Instruct from $13.0\%$ to $84.0\%$. The improvements remain consistent across different background types and hidden character scales, demonstrating its effectiveness.

Beyond the illusion task, our findings suggest that part of the MLLM's vulnerability may stem from a perceptual misalignment between the model and humans, rather than its insufficient knowledge or model capacity. The effectiveness of SMSP further demonstrates that such misalignment can be mitigated through perception-level adjustments prior to inference, without retraining or modifying model parameters.

Our main contributions are as follows:
\begin{itemize}
    \item We construct IlluChar, a comprehensive and challenging visual illusion dataset that reveals the vulnerability of MLLMs to character-based illusion images.
    \item We identify and characterize the high-frequency attention bias as a key underlying mechanism for MLLMs' failure.
    \item We propose SMSP, a human-aligned and plug-and-play strategy that effectively improves model performance on illusion images with all considered background types and hidden pattern scales.
\end{itemize}

\section{Related Work}

\subsection{Visual Illusions}

Many studies have investigated various types of visual illusions, including classical cognitive illusions \cite{zhang2025illusionbench, guan2024hallusionbench, makowski2021parametric, zhang2023grounding, panagopoulou2024evaluating, hirsch2020color}, real scene illusions \cite{shahgir2024illusionvqa}, overlay-style illusions \cite{burgert2024diffusion}, and images with geometric transformations \cite{geng2024visual}.

However, only a limited number of studies investigate the hidden-pattern illusions. Existing works mainly focus on illusion images with simple embedding patterns \cite{rostamkhani2025illusory, ding2025illusioncaptcha, fan2023challenging} or malicious content \cite{qu2025hate}. Compared to prior works, our dataset embeds a more challenging pattern type--characters--and introduces richer variations in both background construction and hidden-character scales, better simulating the complexity of real-world illusion scenarios.

\subsection{Analysis and Mitigation of Hidden-Pattern Visual Illusions}

Existing research provides limited investigation into the underlying causes of MLLM's failure on the illusion task. Although some studies have identified deficiencies in MLLMs' visual perception \cite{anvekar2025perceptual, zhang2023grounding} or analyzed the models' image encoding and attention \cite{qu2025hate}, they have not connected these phenomena with the unique features of the illusion images. As for mitigation methods \cite{rostamkhani2025illusory, qu2025hate}, they are mostly simple and heuristic. Based on these works, we further conduct a more comprehensive analysis into the underlying mechanism of MLLM's failure, and propose a plug-and-play strategy that aligns with humans' perceptual strategies.

\begin{table*}[h]
    \caption{Accuracies (\%) of six MLLMs and human participants on IlluChar.}
    \label{tab:dataset_result}
    \resizebox{\linewidth}{!}{
    \begin{tabular}{l cccc cccc cccc}
    \toprule
    \textbf{Models} &
    \multicolumn{4}{c}{\textbf{Origin}} & 
    \multicolumn{4}{c}{\textbf{Noise}} & 
    \multicolumn{4}{c}{\textbf{Semantic}} \\
    \cmidrule(lr){2-5} \cmidrule(lr){6-9} \cmidrule(lr){10-13}
    & 
    \textbf{Large} & \textbf{Medium} & \textbf{Small} & \textbf{Avg.} &
    \textbf{Large} & \textbf{Medium} & \textbf{Small} & \textbf{Avg.\ ($\Delta\downarrow$)} &
    \textbf{Large} & \textbf{Medium} & \textbf{Small} & \textbf{Avg.\ ($\Delta\downarrow$)} \\
    \midrule
    Qwen3-VL-8B & 87.1 & 99.3 & 97.1 & 93.2 & 0.6 & 11.4 & 48.7 & 16.7\ (-76.5) & 2.4 & 6.4 & 6.7 & 3.8\ (-89.4) \\
    GLM-4.5V & 93.1 & 97.1 & 97.1 & 95.3 & 1.3 & 26.3 & 75.7 & 28.5\ (-66.8) & 3.6 & 6.0 & 13.3 & 4.9\ (-90.4) \\
    Qwen3-VL-235B-A22B & 95.3 & 100.0 & 99.3 & 97.7 & 0.9 & 16.4 & 55.4 & 20.0\ (-77.7) & 2.7 & 7.8 & 6.7 & 4.4\ (-93.3) \\
    GPT-5.2 & 77.2 & 99.3 & 98.6 & 89.1 & 3.1 & 2.9 & 26.0 & 9.3\ \ \ (-79.8) & 4.0 & 1.4 & 4.0 & 3.3\ (-85.8) \\
    Gemini-2.5-Pro & 91.4 & 98.6 & 97.1 & 94.9 & 1.7 & 17.1 & 74.4 & 25.8\ (-69.1) & 4.7 & 4.6 & 12.0 & 5.2\ (-89.7) \\
    Claude-Sonnet-4.5 & 75.0 & 97.9 & 97.9 & 87.5 & 1.4 & 1.7 & 0.3 & 1.2\ \ \ (-86.3) & 3.0 & 3.2 & 9.3 & 3.5\ (-84.0) \\
    \textbf{Human} & \textbf{-} & \textbf{-} & \textbf{-} & \textbf{-} & \textbf{98.9} & \textbf{100.0} & \textbf{100.0} & \textbf{99.5} & \textbf{98.7} & \textbf{99.3} & \textbf{100.0} & \textbf{99.0} \\
    \bottomrule
    \end{tabular}
    }
\end{table*}

\section{IlluChar: A Character-Based Illusion Dataset}
\subsection{Dataset Overview}
\label{sec:dataset_overview}

\begin{figure}[b]
  \centering
  \includegraphics[width=0.85\linewidth]{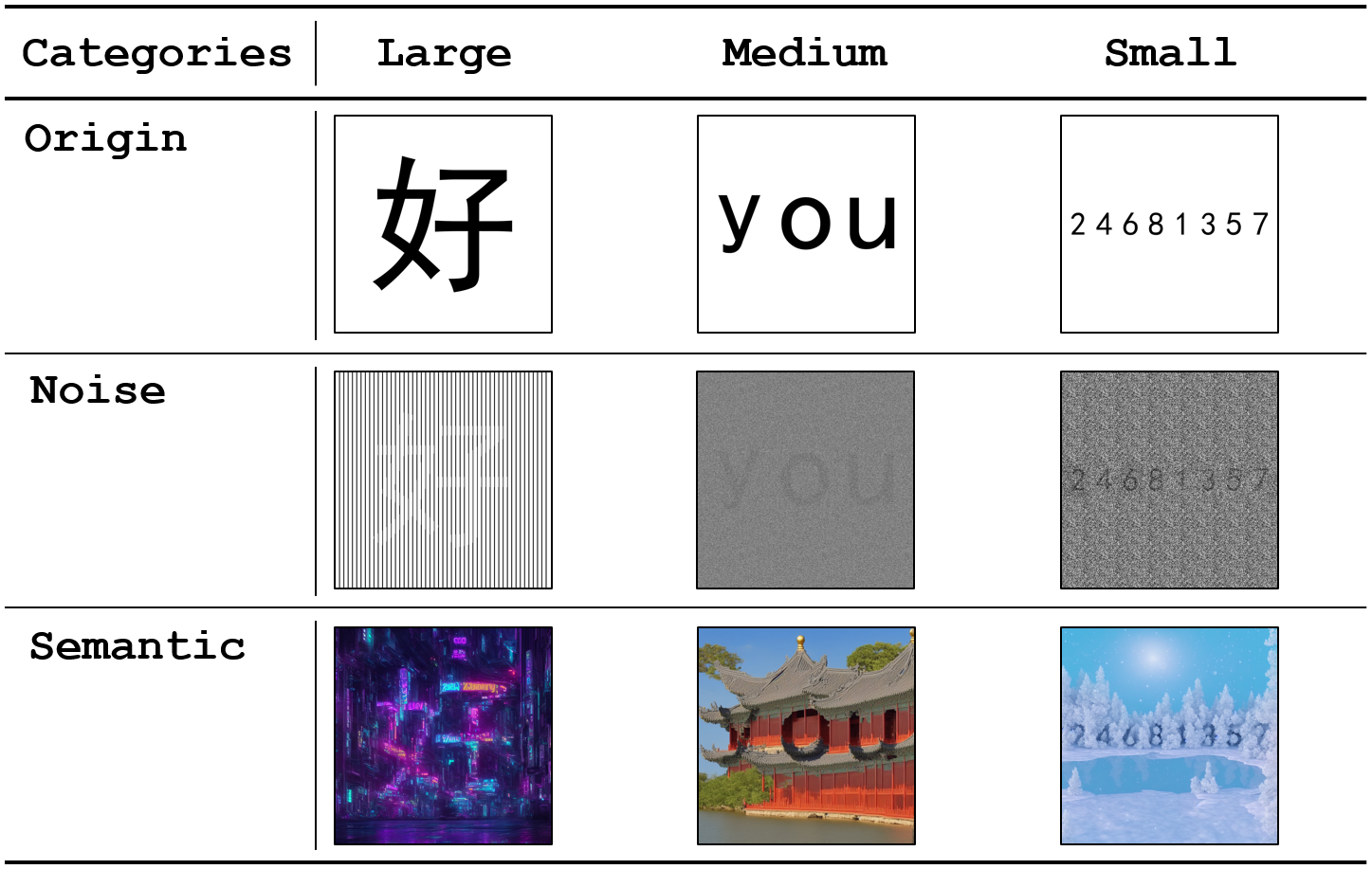}
  \caption{Examples across different categories in IlluChar.}  
  \label{fig:dataset_overview}
  \Description{Some examples of our dataset IlluChar.}
\end{figure}

We introduce \textbf{IlluChar}, a comprehensive and challenging character based illusion dataset. Each image ($1000\times1000$ resolution) embeds specific characters within a generated background.

For hidden patterns, we select commonly used characters, including 10 digits, 52 uppercase and lowercase English letters, and 170 high-usage Chinese characters, covering diverse character structures from simple to complex. Moreover, to simulate real-world settings where the hidden patterns may appear at varying spatial scales, our dataset includes characters of different sizes. Based on their scales, we categorize all the images into \textbf{large, medium, and small} scale groups.

For backgrounds, we consider two categories: \textbf{Semantic Backgrounds} and \textbf{Noise Backgrounds}. Semantic Backgrounds refer to AI-generated realistic scenes, while Noise Backgrounds are non-AI-generated noise textures such as gratings or Gaussian noise. Both of them have been proved effective to generate illusions \cite{rostamkhani2025illusory, fan2023challenging}.

Unlike prior datasets \cite{rostamkhani2025illusory, qu2025hate}, IlluChar mainly focuses on characters, and incorporates a wider range of pattern scales and background types, making it a more comprehensive and challenging benchmark for MLLMs. We provide some examples in Figure \ref{fig:dataset_overview}.




\subsection{Dataset Construction}

We first render the hidden characters of each illusion image onto a clean white background to obtain the corresponding original image. To generate characters of different scales, we place character strings of different lengths within images of the same resolution, thereby producing different character sizes.

We then construct illusion images with two types of backgrounds. For the Semantic Background illusions, we employ Stable Diffusion \cite{rombach2022high} with ControlNet \cite{zhang2023adding} to embed characters into realistic scenes. For the Noise Background illusions, we first generate images filled with a specific noise texture, and then subtly adjust the local texture attributes within the character regions to embed them. More details of IlluChar are provided in the Appendix \ref{sec:appendixB}.



\subsection{Evaluation and Results}
\label{sec:dataset}

We evaluate six representative state-of-the-art MLLMs (Qwen \cite{Qwen3-VL}, GLM \cite{hong2025glm}, GPT \cite{singh2025openai}, Gemini \cite{comanici2025gemini}, and Claude \cite{anthropic2025claude_sonnet45} models), which are widely deployed MLLMs in real-world applications, and cover both open- and closed-source models with diverse architectures. We additionally recruit 10 human participants for comparison.

Results are presented in Table \ref{tab:dataset_result}. We report the recognition accuracy on both original and illusion images, along with the performance drop ($\Delta\downarrow$) from original to illusion images. Our observations are as follows:

(1) \textbf{All models exhibit vulnerabilities to illusion images.} While human participants consistently achieve near-perfect accuracy, all models suffer substantial performance drops (over $65\%$) on illusion images under all settings. Besides, we observe that for most models, the accuracies on Semantic Backgrounds are lower than those on Noise Backgrounds, indicating that they are more vulnerable to realistic scenes.

(2) \textbf{All models demonstrate limited robustness to the scale variations of hidden characters.} As shown in Table \ref{tab:dataset_result}, the model's accuracy decreases as the character size gets larger, suggesting that MLLMs lack sufficient robustness to the scale diversity.


\section{Mechanism: High-Frequency Attention Bias}

\begin{figure}[b]
  \centering
  \includegraphics[width=\linewidth]{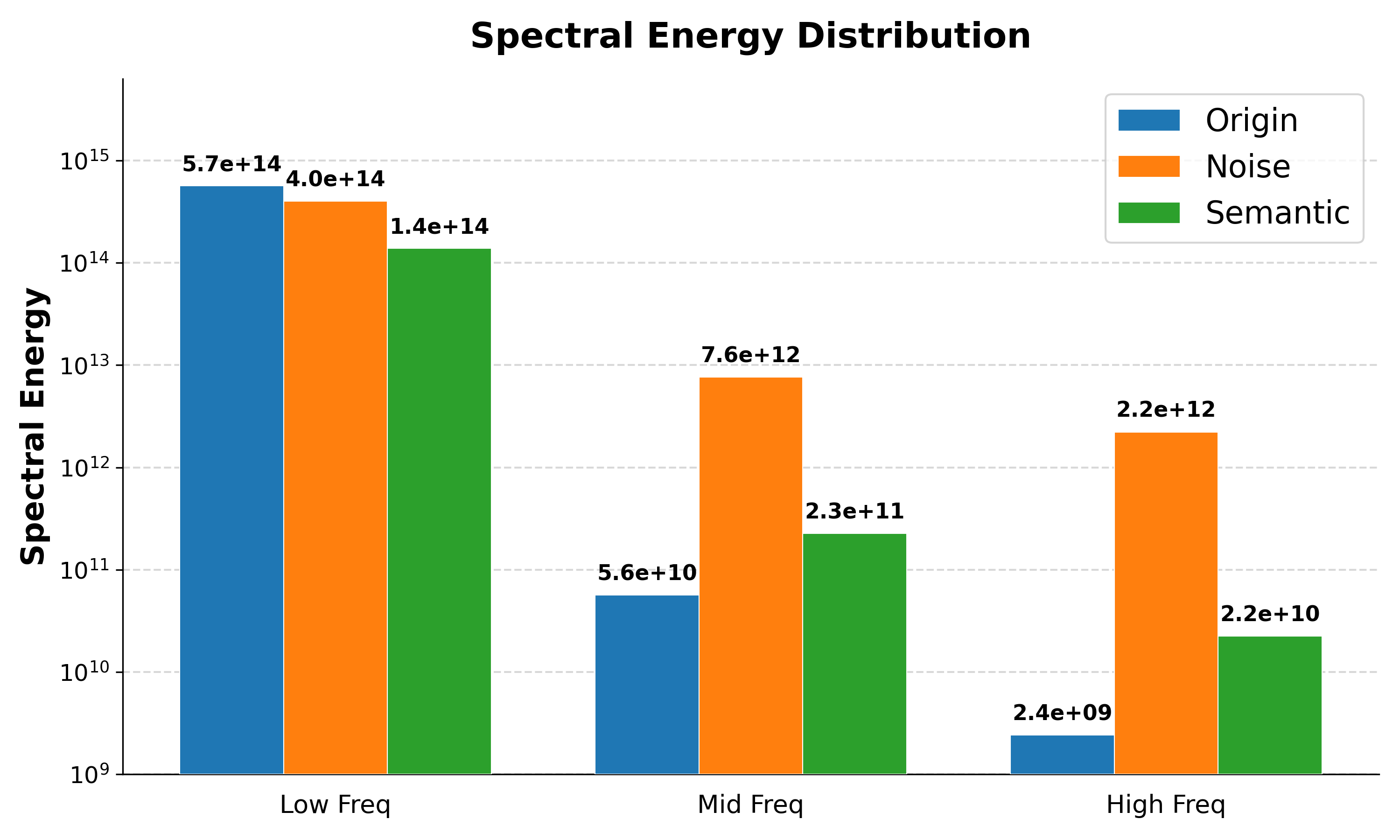}
  \caption{Spectral energy distribution comparison between original and illusion images.}
  \label{fig:frequency_analysis}
  \Description{A bar chart showing the relative spectral energy distribution of three image types.}
\end{figure}

\begin{figure*}
  \centering
  \includegraphics[width=\linewidth]{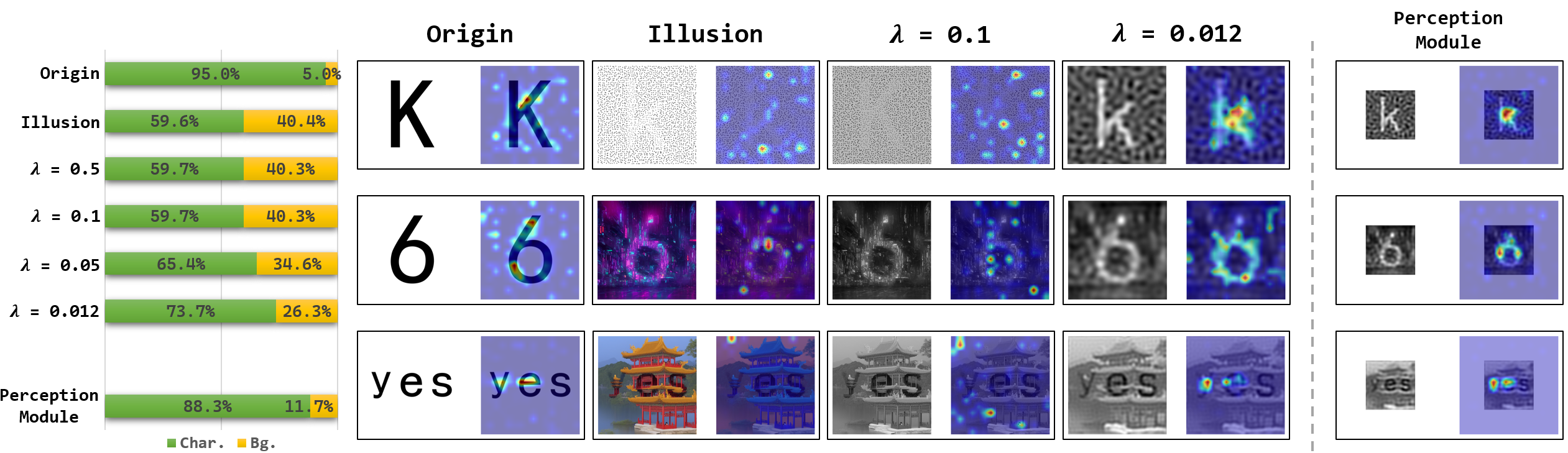}
  \caption{Analysis of the variance in model's attention distribution. \textit{Left:} A quantitative analysis of the model’s high-attention (regions with top $20\%$ attention scores) distribution. \textit{Right:} Three representative examples.}
  \label{fig:attention_exp}
  \Description{Two rows of image examples with corresponding attention relevance maps.}
\end{figure*}

To better understand the MLLM's failure, we further investigate its underlying mechanism by examining the distinctive visual characteristics of illusion images and their impact on the model.

\textbf{RQ1: What are special visual features of illusion images?}

Compared to the original images, illusion images contain backgrounds with richer visual details, which typically correspond to stronger high-frequency components in the frequency domain. To quantify this feature, we analyze their frequency spectra.

Following prior work \cite{ye2024frequency}, we adopt \textbf{spectral energy distribution} to measure the magnitude of different frequency components of an image. Let $F(u,v)$ denote the frequency representation of an image $I\in \mathbb{R}^{H\times W}$, obtained via Fast Fourier Transform (FFT),
\begin{equation}
    F(u,v)=\sum^{H-1}_{x=0}\sum^{W-1}_{y=0}I(x,y)e^{-j2\pi(\frac{ux}{H}+\frac{vy}{W})},
\label{eq:FFT}
\end{equation}
the spectral energy at frequency $r$ is then defined as 
\begin{equation}
    E(r)=\sum_{\sqrt{u^2+v^2}\in[r, r+1)}|F(u,v)|^2.
\end{equation}

For comparison, we aggregate the energy into low- ($r\le100$), mid- ($300\le r\le400$), and high- ($r\ge500$) frequency bands. As shown in Figure \ref{fig:frequency_analysis}, compared to original images, illusion images exhibit significantly higher spectral energy in the middle- and high-frequency bands, regardless of the background types. The result suggests that both types of backgrounds introduce high-frequency signals, while hidden characters are primarily encoded in relatively lower-frequency components.

\textbf{RQ2: How do high-frequency backgrounds affect the model?}

To investigate the impact of the high-frequency signals, we analyze the model's attention distribution. Since most MLLMs adopt the CLIP-ViT-based \cite{radford2021learning} visual encoder \cite{liu2024comprehensive}, we select it as a representative backbone for our analysis, and employ an attention explainability method \cite{chefer2021generic} to obtain attention maps.

Specifically, we measure the proportion of the model's high-attention regions located within character areas versus background areas to show its variance. As shown in Figure \ref{fig:attention_exp}, the model assigns nearly all attention to the character areas in original images, while dropping substantially to below $60\%$ on illusion images, showing a significant attention shift toward the background areas.

We further investigate the cause of this attention shift through a controlled
frequency intervention. Specifically, we keep the low-frequency components unchanged while progressively suppressing high-frequency components by decreasing the filtering radius ratio $\lambda$. As shown in Figure \ref{fig:attention_exp}, stronger high-frequency suppression shifts the model's attention back toward character regions (from $59.6\%$ to $73.7\%$). This indicates that high-frequency signals introduced by the illusion backgrounds drive the attention shift and hinder the recognition of lower-frequency character structures. We term this \textbf{high-frequency attention bias}. Additional analyses on Qwen3-VL-8B and Gemini-2.5-Pro are provided in the Appendix \ref{sec:appendixC}, which show the same phenomenon.

\section{Mitigation Method}

\begin{figure*}[t]
  \centering
  \includegraphics[width=\linewidth]{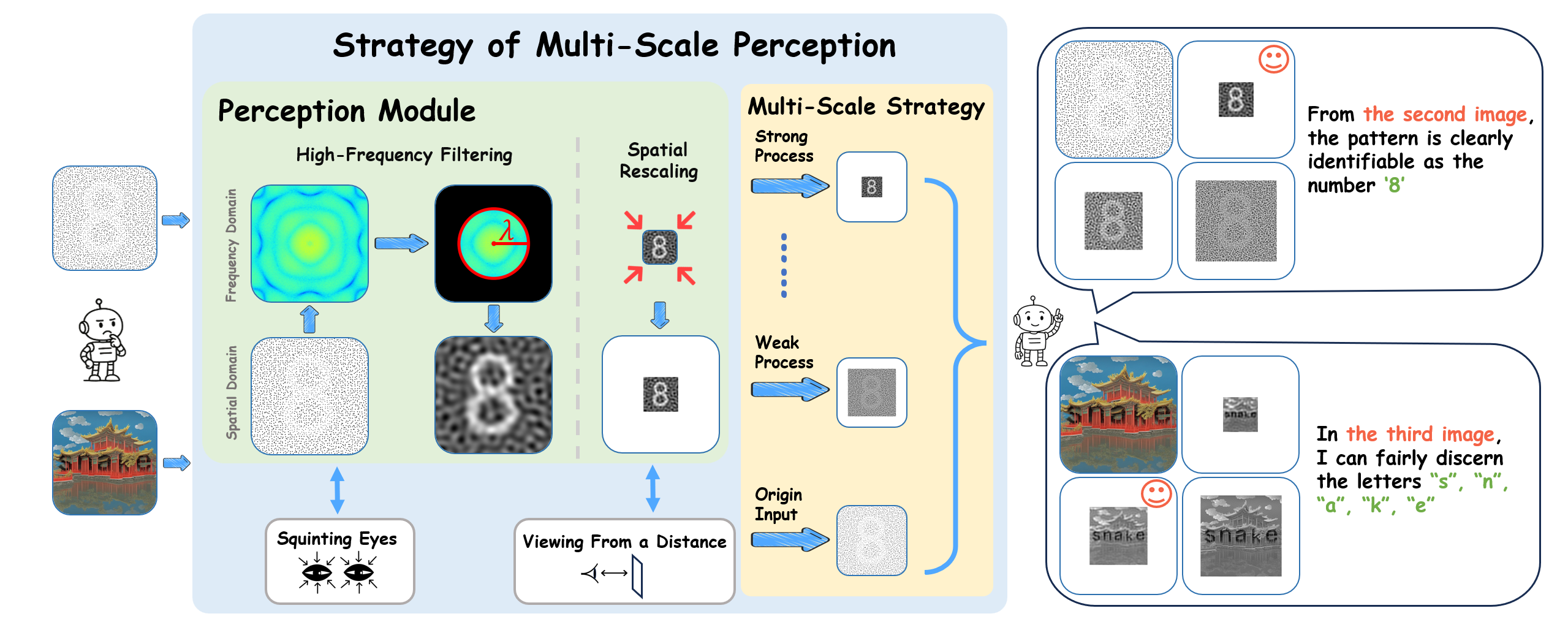}
  \caption{An outline of the Strategy of Multi-scale Perception (SMSP). Two examples are provided to demonstrate the whole process and illustrate how models identify hidden characters with the help of SMSP.}
  \label{fig:SMSP}
  \Description{A diagram showing the Strategy of Multi-scale Perception (SMSP).}
\end{figure*}

\subsection {Inspiration: Human Perceptual Strategies}

When conducting human experiments in Section \ref{sec:dataset}, we observe that 
a similar initial distraction by high-frequency backgrounds can also appear in humans. When encountering illusion images, human participants are initially drawn to the fine-grained backgrounds as well. However, they can instinctively adopt two perceptual adjustment strategies--\textbf{squinting and viewing the image from a distance}--to suppress the distracting backgrounds and obtain a clearer perception, thereby refocusing on hidden patterns. Moreover, humans can dynamically adapt these adjustments to illusions with varying hidden-pattern scales and backgrounds, further enhancing their ability to perceive more diverse images. More details about the human experiments are provided in the Appendix \ref{sec:appendixD}.

In contrast, current MLLMs lack such perceptual strategies. They can only take the original image as the input and cannot spontaneously transform it into a clearer one, making them unable to mitigate the high-frequency attention bias.

\subsection {The Strategy of Multi-Scale Perception}

To bridge this perceptual gap between models and humans' visual capability, we propose \textbf{the Strategy of Multi-Scale Perception (SMSP)}, a plug-and-play framework for MLLMs that simulates the human perceptual strategies. SMSP is designed to achieve three primary objectives:
\begin{itemize}
    \item Attention Recalibration: Mitigating the model's attention bias by introducing a visual perceptual bottleneck to suppress the high-frequency signals.
    \item Scale-Diversity Handling: Consistently improving the model's recognition performance across hidden patterns of varying spatial scales.
    \item Original Capability Preservation: Ensuring minimal degradation of the model's foundational visual reasoning capabilities on standard, out-of-distribution (OOD) tasks.
\end{itemize}

As illustrated in Figure \ref{fig:SMSP}, SMSP consists of two key components, \textbf{Perception Module} and \textbf{Multi-Scale Strategy}, to reach the goals. 

\subsubsection {Perception Module}

The Perception Module acts as a high-frequency signal bottleneck. It simulates two human perceptual adaptation strategies—squinting and viewing the image from a distance—to prune high-frequency details, thereby enabling the model to re-allocate its attention toward hidden content. Specifically, it applies two sequential operations:







\begin{itemize}
    \item \textbf{High-Frequency Filtering}: To simulate human squinting, which biologically filters out high-frequency visual noise, we first transform the input illusion image $I$ into its frequency representation $F(u,v)$ using Equation \ref{eq:FFT}, and then apply a low-pass filter to discard all the noisy high-frequency signals above a specific frequency threshold $\lambda$.
    \item \textbf{Spatial Rescaling}: To simulate human viewing afar, which prioritizes global structure over local textures, we downscale the image with a scale factor $s$, and pad it back to its original resolution with a clean white background. This operation further compresses the background information, and makes the hidden content denser and easier for the model to recognize.
\end{itemize}

Formally, given a perception parameter pair $(\lambda,s)$, the Perception Module processes an input illusion image $I$ into a perception-adjusted variant $\tilde{I}=\mathcal{P}_{(\lambda,s)}(I)$, where $\mathcal{P}$ denotes the Perception Module. The complete processing procedure is outlined in Algorithm \ref{alg:perception_module}.

\begin{algorithm}[t]
\caption{Image processing procedure in the Perception Module}
\label{alg:perception_module}

\KwIn{Input image $I \in \mathbb{R}^{H \times W}$, filtering threshold $\lambda \in (0,1)$, rescaling factor $s \in (0,1)$}
\KwOut{Processed image $\tilde{I}$}

\tcc{\textbf{Operation 1: High-Frequency Filtering}}
$F = \text{FFTShift}(\text{FFT2D}(I))$ \tcp{\textcolor{gray}{Transform the image into frequency domain}}
\For{$(u,v)$ in $(0,H)\times(0,W)$}{
    $F_{\text{filtered}}[u,v] = 
    \begin{cases}
        F[u,v], & \sqrt{(u-\frac{H}{2})^2+(v-\frac{W}{2})^2} \le \min(\frac{H}{2},\frac{W}{2})\cdot \lambda \\
        0, & \text{otherwise}
    \end{cases}$
}\tcp{\textcolor{gray}{Discard all the high-frequency signals}}
$I_{\text{filtered}} = |\text{IFFT2D}(\text{IFFTShift}(F_{\text{filtered}}))|$ \tcp{\textcolor{gray}{Transform back}}

\tcc{\textbf{Operation 2: Spatial Rescaling}}
$I_{\text{scaled}} = \text{Resize}(I_{\text{filtered}}, (H\cdot s, W\cdot s))$ \tcp{\textcolor{gray}{Downscale the filtered image by the scale factor $s$}}
$\tilde{I} = \text{CreateCanvas}(H,W,\text{white})$

$\text{PasteAtCenter}(\tilde{I}, I_{\text{scaled}})$

\Return $\tilde{I}$
\end{algorithm}

\begin{table*}[t]
    \caption{Accuracies (\%) of six MLLMs with different methods on IlluChar.}
    \label{tab:main_results}
    \small
    \resizebox{\linewidth}{!}{
    \begin{tabular}{l | cccc cccc cccc | cccc cccc cccc}
    \toprule
    \textbf{Method} 
    & \multicolumn{4}{c}{\textbf{Origin}} 
    & \multicolumn{4}{c}{\textbf{Noise}}
    & \multicolumn{4}{c}{\textbf{Semantic}}
    & \multicolumn{4}{c}{\textbf{Origin}} 
    & \multicolumn{4}{c}{\textbf{Noise}}
    & \multicolumn{4}{c}{\textbf{Semantic}} \\
    \cmidrule(lr){2-5} \cmidrule(lr){6-9} \cmidrule(lr){10-13}
    \cmidrule(lr){14-17} \cmidrule(lr){18-21} \cmidrule(lr){22-25}
    & L & M & S & Avg. & L & M & S & Avg. & L & M & S & Avg.
    & L & M & S & Avg. & L & M & S & Avg. & L & M & S & Avg. \\
    \midrule
    & \multicolumn{12}{c|}{\colorbox{cyan!5}{\textbf{Qwen3-VL-8B-Instruct}}}
    & \multicolumn{12}{c}{\colorbox{yellow!10}{\textbf{GLM-4.5V}}} \\
    \cmidrule(lr){2-25}
    Vanilla  & 87.1 & 99.3 & 97.1 & 93.2 & 0.6 & 11.4 & 48.7 & 16.7 & 2.4 &  6.4 & 6.7 & 3.8
             & 93.1 & 97.1 & 97.1 & 95.3 & 1.3 & 26.3 & 75.7 & 28.5 & 3.6 & 6.0 & 13.3 & 4.9 \\
    
    CoT      & 86.6 & 99.3 & 97.9 & 93.2 & 1.0 & 12.3 & 49.9 & 17.5 & 1.6 & 8.9 & 6.7 & 3.9
             & 92.2 & 92.9 & 96.4 & 93.6 & 1.5 & 24.3 & 74.7 & 27.7 & 3.3 & 6.7 & 10.7 & 4.7 \\

    Filtered & 94.4 & 99.3 & 97.9 & 96.7 & 20.9 & 59.1 & 17.1 & 30.3 & 26.0 & 81.9 & 80.0 & 44.8
             & 94.0 & 94.3 & 81.4 & 90.6 & 27.3 & 97.1 & 51.7 & 53.1 & 30.5 & 90.8 & 85.3 & 50.5 \\
    
    Blur with Histogram & 90.9 & 95.0 & 10.0 & 69.9 & 42.6 & 66.0 & 1.4 & 37.7 & 41.8 & 42.9 & 6.7 & 39.6
             & 90.5 & 92.9 & 12.1 & 69.7 & 45.4 & 84.7 & 6.0 & 45.4 & 37.1 & 64.5 & 28.0 & 43.8 \\
             
    \textbf{SMSP (Ours)}
             & \textbf{95.3} & \textbf{100.0} & \textbf{97.9} & \textbf{97.3} & \textbf{79.1} & \textbf{95.4} & \textbf{93.0} & \textbf{87.3} & \textbf{74.0} & \textbf{83.7} & \textbf{82.7} & \textbf{77.2}
             & \textbf{97.8} & \textbf{100.0} & \textbf{98.6} & \textbf{98.6} & \textbf{69.9} & \textbf{97.4} & \textbf{92.1} & \textbf{83.5} & \textbf{58.5} & \textbf{92.9} & \textbf{88.0} & \textbf{69.8} \\

    \midrule
    & \multicolumn{12}{c|}{\colorbox{blue!10}{\textbf{Qwen3-VL-235B-A22B-Instruct}}} 
    & \multicolumn{12}{c}{\colorbox{purple!10}{\textbf{Claude-Sonnet-4.5}}}\\
    \cmidrule(lr){2-25}
    Vanilla  & 95.3 & 100.0 & 99.3 & 97.7 & 0.9 & 16.4 & 55.4 & 20.0 & 2.7 & 7.8 & 6.7 & 4.4 
             & 75.0 & 97.9 & 97.9 & 87.5 & 1.4 & 1.7 & 0.3 & 1.2 & 3.0 & 3.2 & 9.3 & 3.5 \\
    
    CoT      & 94.8 & 100.0 & 100.0 & 97.7 & 1.7 & 15.9 & 52.9 & 19.6 & 3.0 & 6.4 & 9.3 & 4.4
             & 75.0 & 97.9 & 98.6 & 87.7 & 1.5 & 1.9 & 0.0 & 1.2 & 5.3 & 3.9 & 9.3 & 5.2 \\
    
    Filtered & 97.4 & 100.0 & 96.4 & 97.9 & 24.4 & 73.1 & 28.6 & 38.9 & 33.9 & 88.3 & 93.3 & 52.7 
             & 79.3 & 97.9 & 65.7 & 80.7 & 6.4 & 28.0 & 2.6 & 11.2 & 9.1 & 46.8 & 49.3 & 22.0 \\
    
    Blur with Histogram & 94.8 & 99.3 & 21.4 & 76.0 & 42.9 & 80.0 & 3.1 & 42.2 & 59.5 & 60.6 & 17.3 & 56.8
             & 59.5 & 57.9 & 0.7 & 43.0 & 25.8 & 60.0 & 3.6 & 29.1 & 7.0 & 13.1 & 0.0 & 8.2 \\
    
    \textbf{SMSP (Ours)} 
             & \textbf{99.6} & \textbf{100.0} & \textbf{100.0} & \textbf{99.8} & \textbf{79.7} & \textbf{98.0} & \textbf{94.0} & \textbf{88.6} & \textbf{78.2} & \textbf{89.7} & \textbf{94.7} & \textbf{82.4}
             & \textbf{86.6} & \textbf{99.3} & \textbf{98.6} & \textbf{93.4} & \textbf{41.3} & \textbf{80.6} & \textbf{65.4} & \textbf{58.6} & \textbf{32.2} & \textbf{61.0} & \textbf{60.0} & \textbf{41.9} \\
    
    \midrule
    & \multicolumn{12}{c|}{\colorbox{orange!10}{\textbf{GPT-5.2}}} 
    & \multicolumn{12}{c}{\colorbox{green!10}{\textbf{Gemini-2.5-Pro}}} \\
    \cmidrule(lr){2-25}
    Vanilla  & 77.2 & 99.3 & 98.6 & 89.1 & 3.1 & 2.9 & 26.0 & 9.3 & 4.0 & 1.4 & 4.0 & 3.3
             & 91.4 & 98.6 & 97.1 & 94.9 & 1.7 & 17.1 & 74.4 & 25.8 & 4.7 & 4.6 & 12.0 & 5.2 \\
    
    CoT      & 77.2 & 97.9 & \textbf{99.3} & 88.9 & 3.3 & 2.9 & 26.7 & 9.6 & 5.3 & 1.4 & 4.0 & 4.2
             & 92.2 & 98.6 & 97.9 & 95.5 & 1.2 & 17.0 & 68.9 & 24.0 & 4.3 & 8.5 & 13.3 & 6.1 \\
    
    Filtered & 75.4 & 96.4 & 67.9 & 79.1 & 13.9 & 64.6 & 33.4 & 33.1 & 13.8 & 66.3 & \textbf{85.3} & 33.0
             & \textbf{97.4} & \textbf{100.0} & 87.9 & 95.5 & 26.5 & 85.6 & 43.7 & 47.3 & 36.1 & 77.0 & 74.7 & 49.8 \\
    
    Blur with Histogram & 69.4 & 59.3 & 0.7 & 47.9 & 33.4 & 58.7 & 2.9 & 32.0 & 23.1 & 24.8 & 5.3 & 22.3
             & 88.8 & 98.6 & 97.9 & 93.9 & 45.2 & 18.9 & 68.9 & 44.5 & 36.2 & 11.0 & 13.3 & 27.8 \\
    
    \textbf{SMSP (Ours)} 
             & \textbf{87.9} & \textbf{99.3} & 98.6 & \textbf{93.9} & \textbf{44.8} & \textbf{85.1} & \textbf{70.6} & \textbf{62.9} & \textbf{38.4} & \textbf{72.3} & 77.3 & \textbf{50.2}
             & 92.7 & 99.3 & \textbf{97.9} & \textbf{95.9} & \textbf{55.9} & \textbf{91.6} & \textbf{85.6} & \textbf{73.8} & \textbf{45.8} & \textbf{78.0} & \textbf{76.0} & \textbf{56.6} \\
    
    \bottomrule
    \end{tabular}
    }
\end{table*}


To preliminarily validate its effectiveness, we again trace changes in the model's attention distribution. As shown in Figure \ref{fig:attention_exp}, applying the Perception Module can effectively make the image clearer and restore the model's focus from $59.6\%$ to $88.3\%$, suggesting that the module can mitigate the high-frequency attention bias, guiding the model to perceive illusions as a human would.

\subsubsection {Multi-Scale Strategy}

We further employ the Multi-Scale Strategy to address the scale diversity and preserve the model's original performance on OOD tasks.

First, the Multi-Scale Strategy generates $K$ different perceptually processed variants $\tilde{I}_i$ using the Perception Module, each with a distinct parameter pair $(\lambda_i, s_i)$. For the variant with the strongest processing strength (smallest $\lambda$ and $s$), it experiences the most aggressive filtering and downscaling, making it effective to reveal large-scale patterns. Conversely, milder variants (larger $\lambda$ and $s$) are suited for smaller patterns to avoid blurring out hidden contents. We arrange these variants by their processing strength, from strong to weak:

\begin{equation}
\lambda_1 < \lambda_2 < \dots < \lambda_K, \quad s_1 < s_2 < \dots < s_K.
\end{equation}

Subsequently, inspired by the multi-scale pyramid paradigm in classical image processing \cite{adelson1984pyramid}, we formulate these perception parameters as a geometric progression. By first empirically anchoring the boundary parameters $(\lambda_1, s_1)$ and $(\lambda_K, s_K)$, we then interpolate other parameters geometrically:
\begin{equation}
\lambda_i = \lambda_1 \cdot \left(\frac{\lambda_K}{\lambda_1}\right)^{\frac{i-1}{K-1}}, \quad s_i = s_1 \cdot \left(\frac{s_K}{s_1}\right)^{\frac{i-1}{K-1}}, \quad i = 2, \dots, K-1.
\label{eq:geometric_params}
\end{equation}

We further guarantee the model's original capabilities by incorporating the original image $I$ as an additional input, compensating for any original visual details lost during the perceptual operations.

Ultimately, SMSP generates an input tuple with $K+1$ images:
\begin{equation}
  \text{I}_{\text{SMSP}} = (I, \tilde{I_1}, \tilde{I_2},...,\tilde{I_K}),
\end{equation}
where $\tilde{I_i}=\mathcal{P}_{(\lambda_i,s_i)}(I)$ denotes the $i$-th perceptually adjusted variant. By jointly feeding these images into MLLMs, SMSP enables the model to access and focus on the most informative variant, which aligns with humans' ability to dynamically adjust perception for different images.

\section{Experiments}
\label{sec:experiments}

\subsection{Settings}

\paragraph{Dataset \& Models}

We evaluate the performance of six MLLMs on IlluChar to perform our study. The evaluated models are the same in Section \ref{sec:dataset}, including Qwen3-VL-8B-Instruct (235B) \cite{Qwen3-VL}, GLM-4.5V \cite{hong2025glm}, GPT-5.2 \cite{singh2025openai}, Gemini-2.5-Pro \cite{comanici2025gemini}, and Claude-Sonnet-4.5 \cite{anthropic2025claude_sonnet45}.

\paragraph{Baselines}

Due to the limited number of effective methods for visual illusions, we compare our method against four baselines: (1) \textbf{Vanilla}: The original models without any additional processing or prompting. (2) \textbf{Chain-of-Thought Prompting (CoT)} \cite{wei2022chain}: We use a CoT prompting technique to guide models to imagine visualizing the image using the proposed perceptual strategies. (3) \textbf{Filtered} \cite{rostamkhani2025illusory}: a sequence of Gaussian blurs is applied to the image, followed by an additional sharpening operation. (4) \textbf{Blur with Histogram} \cite{qu2025hate}: it first applies a blur operation, and then equalizes the histogram of image luminance. Additional comparisons with image-processing baselines are provided in the Appendix \ref{sec:appendixG}.

\paragraph{Evaluation}

We evaluate the recognition accuracy of MLLMs on both original and illusion character images. We employ a hybrid evaluation method that combines both strict string-matching and GPT-instructed techniques (more details are provided in Appendix \ref{sec:appendixE}). To assess its reliability, we manually verify the evaluation results on 1000 randomly sampled instances, with only 5 errors, demonstrating its high reliability.

\paragraph{SMSP Parameter Settings}

To effectively cover various scales of hidden contents, we set the variant number to $K=3$. For perception parameters, we first determine the boundary parameters $(\lambda_1, s_1)=(0.012,0.1)$ and $(\lambda_K, s_K)=(0.05,0.4)$ using small validation subsets, and then derive other parameters via the geometric interpolation in Equation \ref{eq:geometric_params}. A detailed analysis of the parameter selection is provided in the Appendix \ref{sec:appendixI}.



\subsection{Main Results}

We report the results in Table \ref{tab:main_results}. Our findings are as follows:

\textbf{(1) SMSP consistently improves performance on illusion images across all models and background types.} As shown in Table \ref{tab:main_results}, SMSP substantially boosts accuracies on illusion images for all evaluated models, demonstrating its strong transferability. Moreover, the significant improvements are observed on both Noise and Semantic background illusions, showing its effectiveness across different illusion backgrounds. For example, the average accuracy of Qwen3-VL-8B-Instruct significantly increases from $16.7\%$ and $3.8\%$ to $87.3\%$ and $77.2\%$ on Noise and Semantic backgrounds, respectively. Averaged over all illusion images (Noise + Semantic), its overall accuracy improves from $13.0\%$ to $84.0\%$, significantly outperforming baseline methods.

\textbf{(2) SMSP consistently improves performance across all hidden character scales.} While baseline methods typically achieve improvements on illusions with partial character scales, their gains remain limited on others. In contrast, SMSP consistently improves accuracies across all the scales, demonstrating its effectiveness in handling the scale diversity of hidden characters.

\textbf{(3) SMSP preserves—and often enhances—accuracies on original clean inputs.} Unlike baseline methods that often degrade model performance on already high-accuracy original images, SMSP improves illusion accuracy while maintaining and often slightly improving performance on clean inputs, demonstrating its robustness.



\subsection{Ablation Study}
\label{sec:ablation}



\paragraph{Effect of SMSP Components} 

We perform ablation studies on Qwen3-VL-8B-Instruct to isolate the contributions of each component. Specifically, we examine the following configurations: (1) the complete \textbf{SMSP} framework; (2) \textbf{w/o High-Frequency Filtering} and \textbf{w/o Spatial Rescaling}, which remove each operation from the Perception Module, respectively; and (3) \textbf{Single Processed Variant ($i = 1, 2, 3$)}, which retains only one processed variant together with the original image, where $i$ denotes the selected variant. We report the average accuracy on original images, and scale-wise results averaged on illusion images.

As shown in Table \ref{tab:ablation_study}, removing either High-Frequency Filtering or Spatial Rescaling leads to a noticeable performance degradation, indicating that the two operations provide complementary benefits and are both indispensable. Moreover, using only a single perceptual variant results in limited performance across all hidden pattern scales. Each performs well only within a narrow range of scales, while failing to generalize to others. The results confirm the necessity of the Multi-Scale Strategy, which can effectively mitigate the limitation of using a fixed processing strength.


\begin{table}[h]
    \caption{Ablation study on Qwen3-VL-8B-Instruct.}
    \label{tab:ablation_study}
    \resizebox{0.9\linewidth}{!}{
    \begin{tabular}{l | c ccc}
    \toprule
    \textbf{Method} & \textbf{Origin} & \multicolumn{3}{c}{\textbf{Illusion}} \\
    \cmidrule(lr){3-5}
    & & Large & Medium & Small \\
    \midrule
    \textbf{SMSP} & 97.3 & \textbf{77.2} & \textbf{92.1} & \textbf{92.0} \\
    \midrule
    \multicolumn{5}{c}{\textit{\textbf{Analysis of the Perception Module}}} \\
    \midrule
    w/o Low-pass Filtering & 96.3 & 31.8 & 84.3 & 87.9 \\
    w/o Resizing with Padding & 95.5 & 42.8 & 88.9 & 91.2 \\
    \midrule
    \multicolumn{5}{c}{\textit{\textbf{Analysis of the Multi-Scale Strategy}}} \\
    \midrule
    Single Processed Variant ($i=1$) & 96.1 & 76.1 & 65.4 & 47.0  \\
    Single Processed Variant ($i=2$) & \textbf{98.0} & 54.3 & 90.5 & 74.7 \\
    Single Processed Variant ($i=3$) & 95.9 & 15.1 & 87.2 & 88.0 \\
    \bottomrule
    \end{tabular}
    }
\end{table}




\paragraph{Discussion on Processed Variant Number $K$}

\begin{figure}
  \centering
  \includegraphics[width=\linewidth]{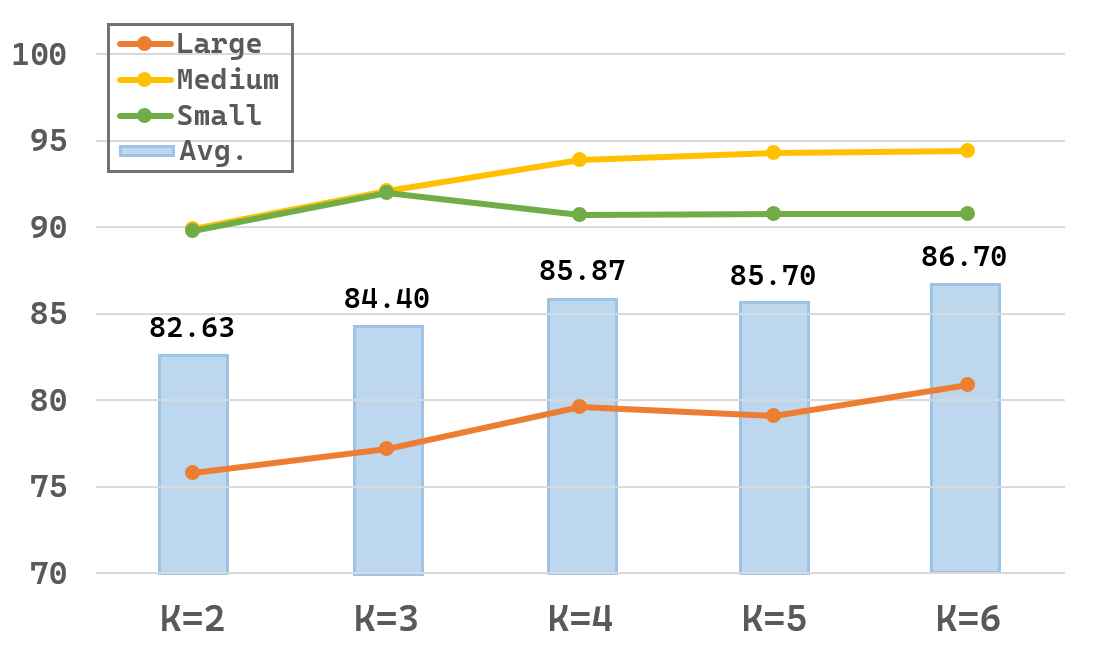}
  \caption{Comparison of the model's accuracy (\%) on illusion images across different hidden pattern scale ranges for varying processed variant numbers $K$.}
  \label{fig:ablation_k}
  \Description{A performance comparison showing model accuracy across different hidden pattern scale ranges under varying branch numbers K.}
\end{figure}

We further analyze the impact of the variant number $K$ on both recognition accuracy and computational cost. For each $K$, we fix the same boundary parameters and derive the remaining parameters via Equation \ref{eq:geometric_params}. As shown in Figure \ref{fig:ablation_k}, all tested values of $K$ consistently and effectively improve performance across various hidden pattern scale ranges (above $75\%$). The overall accuracy generally increases as $K$ grows, indicating that additional variants enhance scale coverage. However, when $K$ becomes excessively large (e.g., $K \geq 4$), the performance gain becomes marginal, suggesting diminishing returns from further increasing the number of variants.

Furthermore, we quantify the computational cost introduced by SMSP. As shown in Table \ref{tab:cost}, compared to the vanilla model, SMSP increases the number of input tokens and runtime, both of which grow with the variant number $K$. Balancing performance and efficiency, we adopt $K = 3$ as the default setting, which achieves $84.3\%$ accuracy while moderately increasing the runtime from $1.08$s to $1.43$s ($1.32\times$), showing a favorable trade-off. The results further indicate that despite introducing additional input tokens, SMSP maintains competitive inference efficiency, which aligns with humans' rapid perceptual adjustments to illusions, instead of additional knowledge acquisition or training.

\begin{table}
    \caption{Comparison of per-sample computational costs across different variant numbers $K$ and the vanilla setting.}
    \label{tab:cost}
    
    \resizebox{0.85\linewidth}{!}{
    \begin{tabular}{l | c c c c c c}
    \toprule
    \textbf{Cost} &  \textbf{Vanilla} & \textbf{K=2} & \textbf{K=3} & \textbf{K=4} & \textbf{K=5} & \textbf{K=6} \\
    \midrule
    Input Tokens & 1023 & 1986 & 2949 & 3912 & 4875 & 5838 \\
    Runtime (s) & 1.08 & 1.30 & 1.43 & 1.77 & 1.92 & 2.07 \\
    \bottomrule
    \end{tabular}
    }
\end{table}

\subsection{Generalization Ability}

To assess whether SMSP can generalize beyond character-based illusions, we evaluate its performance on illusions embedding other types of patterns \cite{rostamkhani2025illusory}, such as animals \cite{lin2024sdxl}, FashionMNIST \cite{xiao2017fashion} and MNIST \cite{lecun2002gradient}. Inspired by the HatefulIllusion dataset \cite{qu2025hate}, we also evaluate our method on detecting harmful patterns hidden in illusion images. As shown in Table \ref{tab:other_dataset}, SMSP consistently achieves the highest accuracy on illusion images of all hidden pattern types, demonstrating its strong generalization ability.

\begin{table}[h]
    \caption{Accuracies (\%) of different methods on illusion images with different hidden pattern types.}
    \label{tab:other_dataset}

    \resizebox{\linewidth}{!}{
    \begin{tabular}{l | c c c c}
    \toprule
    \textbf{Method} & \textbf{Animals} & \textbf{Fashion-} & \textbf{MNIST} & \textbf{Harmful} \\
    & & \textbf{MNIST} & & \textbf{Pattern} \\
    \midrule
    Vanilla & 34.0 & 8.5 & 22.0 & 4.3 \\
    Filtered & 93.5 & 50.5 & 71.5 & 66.8 \\
    Blur with Histogram & 71.0 & 29.0 & 76.0 & 63.5 \\
    \midrule
    \textbf{SMSP} & \textbf{97.5} & \textbf{51.5} & \textbf{88.0} & \textbf{71.0} \\
    \bottomrule
    \end{tabular}
    }
\end{table}

\subsection{General Capability Preservation}
To ensure that SMSP is safe for general deployment, we evaluate its compatibility with three standard Visual Question Answering (VQA) tasks: SimpleVQA \cite{cheng2025simplevqa}, MMStar \cite{chen2024we}, and RealWorldQA \cite{realworldqa2024}. As reported in Table \ref{tab:non_illusion_task}, SMSP maintains comparable model performance with only minor changes, suggesting it does not interfere with the model's original visual reasoning capabilities. To further isolate the source of this robustness, we also evaluate SMSP without the original input branch. The results show a substantial performance drop, indicating that the inclusion of the original image is essential and enough for safeguarding the model’s standard visual reasoning performance.

\begin{table}[h]
    \caption{Accuracies (\%) of different methods on standard VQA tasks, with the model's \textcolor{gray}{original performances} as a reference.}
    \label{tab:non_illusion_task}
    \resizebox{\linewidth}{!}{
    \begin{tabular}{l | c c c}
    \toprule
    \textbf{Method} & \textbf{SimpleVQA} & \textbf{MMStar} & \textbf{RealWorldQA}\\
    \midrule
    \textcolor{gray}{Vanilla} & \textcolor{gray}{47.5} & \textcolor{gray}{71.0} & \textcolor{gray}{70.0}\\
    \midrule
    Filtered & 14.5 & 26.5 & 45.5 \\
    Blur with Histogram & 9.5 & 24.0 & 45.5  \\
    \midrule
    \textbf{SMSP} & \textbf{46.5} & \textbf{68.0} & \textbf{71.0} \\
    \quad w/o Original Input & 5.0 & 25.5 & 44.0 \\
    \bottomrule
    \end{tabular}
    }
\end{table}

\subsection{Case Study}
\label{sec:case_study}

\begin{figure}[h]
  \centering
  \includegraphics[width=\linewidth]{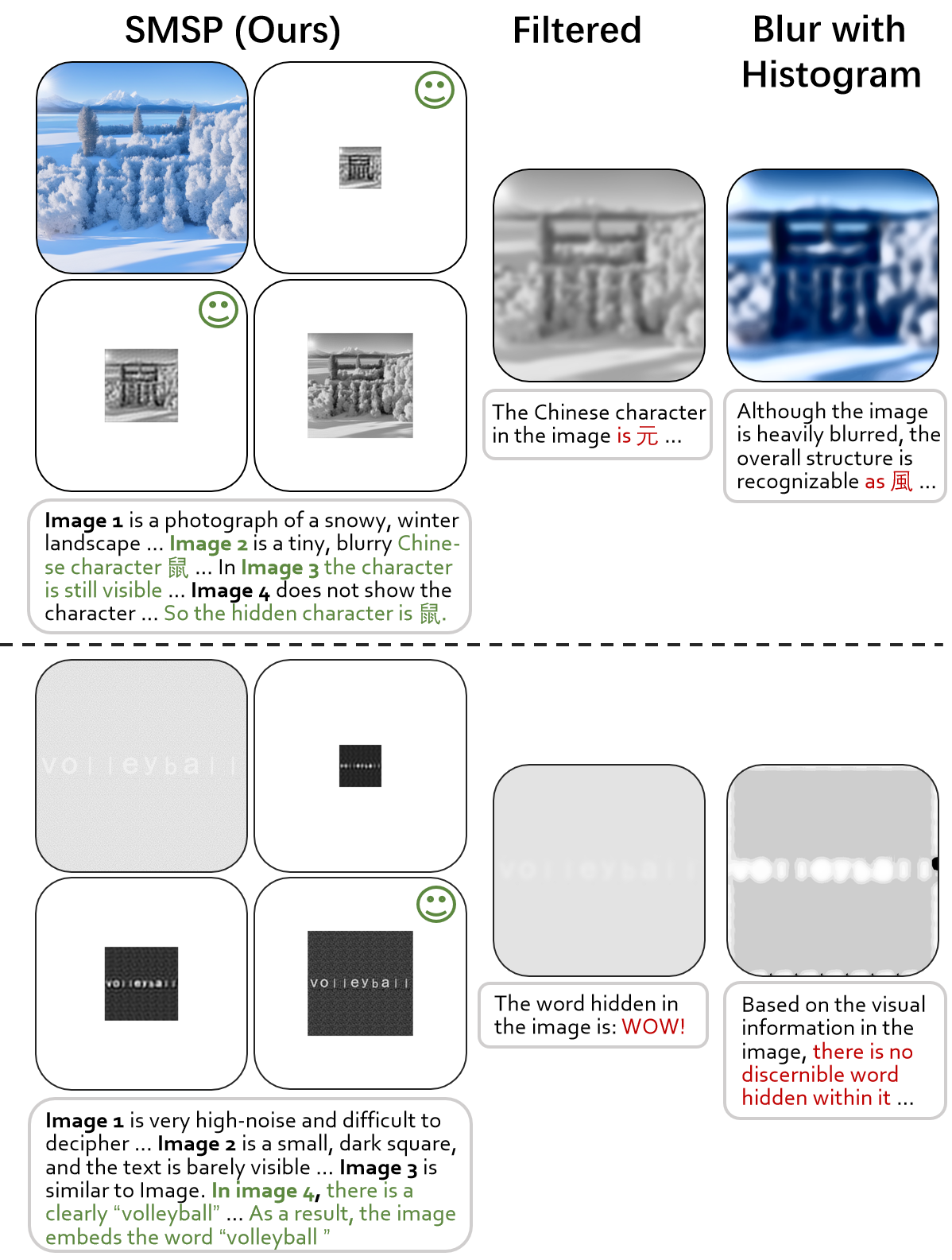}
  \caption{Case study on Qwen3-VL-8B-Instruct.}
  \label{fig:case_study}
  \Description{Examples showing illusion images with hidden characters at different scales, their processed variants generated by SMSP, and corresponding model outputs.}
\end{figure}

We present two representative examples in Figure \ref{fig:case_study} to illustrate the effect of SMSP.

In the first example, a large and complex Chinese character is embedded within a realistic scene. For the Filtered and the Blur with Histogram methods, they fail to effectively remove background distractions and preserve the character's structure, making it difficult for the model to recognize. In contrast, SMSP provides clearer perception-adjusted variants (the second and third images). By jointly considering the four input images, the model can successfully identify the character from strongly processed variants. The second example conceals a small English word in a noise texture, and both baseline methods excessively blur the image. In this example, SMSP enables the model to correctly identify the word through the weakest processed variant (the fourth image), which clearly reveals the answer.

The examples demonstrate that four images constructed by SMSP are necessary and helpful for MLLMs to perceive illusions with different backgrounds and hidden scales. These variants successfully simulate images perceived by humans under their perceptual adjustments, enabling the model to access clearer and more informative visual representations.




\section{Conclusion}

In this work, we systematically investigate MLLMs' vulnerability to hidden-pattern visual illusions. We construct a comprehensive and challenging dataset, IlluChar, and further identify a key failure mechanism, high-frequency attention bias, where models are distracted by high-frequency background information and fail to focus on the lower-frequency hidden content. Based on this mechanism, we propose SMSP, a plug-and-play framework to help models perceive illusion images by simulating human perceptual strategies. Experimental results demonstrate that SMSP consistently and significantly improves performance across all evaluated models on illusion images. For example, it boosts the accuracy of Qwen3-VL-8B-Instruct from $13.0\%$ to $84.0\%$ while maintaining comparable performance on standard VQA benchmarks, indicating that SMSP does not degrade models' foundational visual reasoning capabilities. Our work provides insights into the perceptual gap between humans and multimodal models, suggesting that certain visual failures may stem from models' visual perceptual deficiencies, rather than insufficient knowledge or limited model capacity. Furthermore, the success of SMSP indicates that the perceptual gap can be mitigated through training-free methods without costly retraining. We hope our findings inspire future research on perception-aware multimodal modeling to develop more reliable and human-aligned vision-language systems.

\begin{acks}
This work was supported by the National Natural Science Foundation of China (No. 62506203), and Fundamental and Interdisciplinary Disciplines Breakthrough Plan of the Ministry of Education of China (No. JYB2025XDXM202).
\end{acks}

\bibliographystyle{ACM-Reference-Format}
\bibliography{reference}


\newpage
\appendix

\section{Pilot Study on Visual Illusion Images with Different Hidden Patterns}
\label{sec:appendixA}

We conduct a pilot study to compare the impact of visual illusion images with different hidden patterns on MLLMs. Specifically, we consider several types of hidden patterns. The first group consists of simple patterns, including animals, MNIST digits, and Fashion-MNIST, which are used in the IllusoryVQA \cite{rostamkhani2025illusory} dataset. The second group consists of characters, including digits, English letters, and Chinese characters, which are commonly embedded in illusion images shared on social media. Following \cite{qu2025hate}, we also consider visual illusions that hide harmful or sensitive patterns.

For illusion images containing simple and harmful patterns, we adopt images from IllusoryVQA and HatefulIllusion \cite{qu2025hate} datasets, respectively. For character-based patterns, since limited existing datasets match our requirements, we construct a small set of additional samples by embedding a single large character into similar realistic scenes, ensuring that the data setting remains comparable to other illusion images.

\begin{figure}[h]
  \centering
  \includegraphics[width=\linewidth]{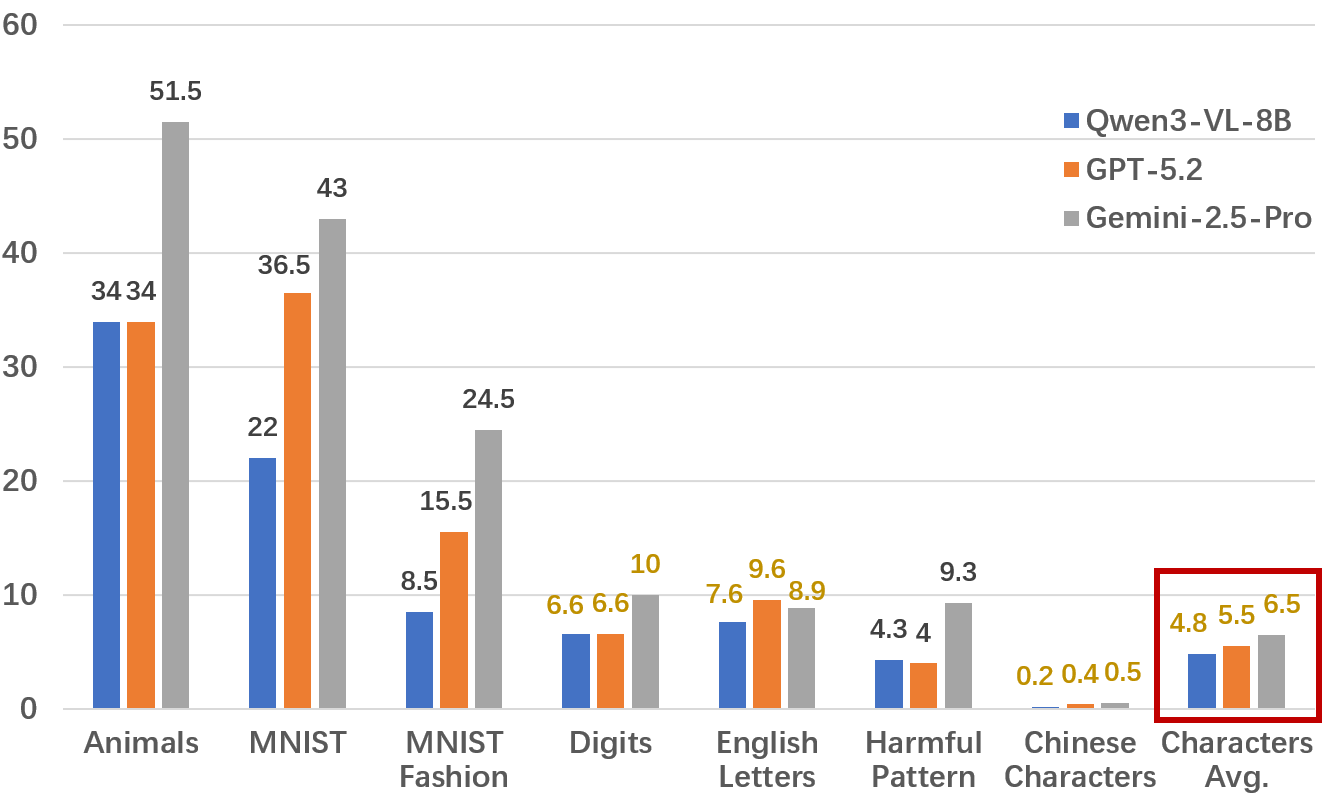}
  \caption{Accuracies (\%) on illusion images with different hidden patterns.}
  \label{fig:pilot_study}
  \Description{Examples showing illusion images with different backgrounds.}
\end{figure}

We evaluate three representative MLLMs on these images. As shown in Figure \ref{fig:pilot_study}, all models achieve significantly lower accuracy on character-based illusions than on images with simple patterns, and the performance is comparable to that with the more complex harmful patterns. In particular, the accuracy on Chinese character illusions is close to $0\%$. The results suggest that embedding characters into illusion images poses a great challenge for current MLLMs. This observation further motivates us to construct the comprehensive character-based dataset, IlluChar.

\section{Dataset Details}
\label{sec:appendixB}

\subsection{Details on Dataset Taxonomy}

In Section \ref{sec:dataset_overview}, we provide a brief overview of the IlluChar dataset. In this section, we present a more detailed description. Based on the type of hidden patterns, IlluChar contains three types of characters:

\begin{itemize}
    \item \textbf{Digits}, including numbers from 0 to 9.
    \item \textbf{English letters}, including uppercase letters A--Z and lowercase letters a--z.
    \item \textbf{Chinese characters}, consisting of 170 commonly used characters. Let $s$ denote the number of strokes in a character. We incorporate 85 structurally simple characters ($3 \le s \le 9$) and 85 structurally complex characters ($s \ge 13$).
\end{itemize}

Based on the spatial scale of hidden characters, IlluChar can also be categorized into Large, Medium, and Small groups. For each image with a resolution of $1000 \times 1000$, suppose the hidden character occupies a region of size $C_H \times C_W$. We categorize images according to the following criteria:

\begin{itemize}
    \item \textbf{Large}: $\max(C_H, C_W) \ge 600$, where each character occupies a large portion of the image.
    \item \textbf{Medium}: $200 \le \max(C_H, C_W) \le 500$, where each character occupies a moderate region of the image.
    \item \textbf{Small}: $\max(C_H, C_W) \le 150$, where the character is relatively small but still clearly recognizable.
\end{itemize}

Based on the background type of illusion images, IlluChar includes two categories: \textbf{Semantic Backgrounds} and \textbf{Noise Backgrounds}. For Semantic Backgrounds, many prior works \cite{ding2025illusioncaptcha, rostamkhani2025illusory, qu2025hate} have already leveraged this type of background to generate illusions. In our dataset, we further empirically select three thematic realistic scenes as Semantic Backgrounds: \textit{Traditional Chinese Architecture (TC)}, \textit{Cyberpunk City (CC)}, and \textit{Winter Valley (WV)}.

For Noise Backgrounds, we design five types of noise textures: \textit{Vertical Gratings (VG)}, \textit{Gaussian Noise (GN)}, \textit{Halftone Dots (HD)}, \textit{Labyrinth Noise (LN)}, and \textit{Micro-text Noise (MN)}, which are empirically observed to produce strong visual illusion effects.

In Table \ref{tab:dataset_num}, we present the number of samples in each category.

\begin{table}[h]
    \caption{Number of samples in each category.}
    \label{tab:dataset_num}
    \resizebox{0.95\linewidth}{!}{
    \begin{tabular}{l | c ccc ccccc}
    \toprule
    \textbf{Categories} & \textbf{Origin} & \multicolumn{3}{c}{\textbf{Semantic}} & \multicolumn{5}{c}{\textbf{Noise}} \\
    \cmidrule(lr){3-5} \cmidrule(lr){6-10}
    & & TC & CC & WV & VG & GN & HD & LN & MN \\
    \midrule
    \textbf{Large} & 232 & 232 & 232 & 232 & 232 & 232 & 232 & 232 & 232 \\
    \textbf{Medium} & 140 & 98 & 95 & 89 & 140 & 140 & 140 & 140 & 140 \\
    \textbf{Small} & 140 & 28 & 26 & 21 & 140 & 140 & 140 & 140 & 140 \\
    \bottomrule
    \end{tabular}
    }
\end{table}

\subsection{Details on Dataset Construction}

In this section, we provide a detailed description of the construction process for illusion images.

For illusions with Semantic Backgrounds, we use Stable Diffusion v1.5 \footnote{The download link to the model: \url{https://huggingface.co/stable-diffusion-v1-5/stable-diffusion-v1-5}} with a variant of ControlNet \footnote{The download link to the model: \url{https://huggingface.co/monster-labs/control_v1p_sd15_qrcode_monster}} to embed characters from the original images (black characters rendered on a white background) into realistic scenes. Empirically, we set the guidance scale to 9 and the number of inference steps to 50 to obtain high-quality illusion images.

For Noise Background illusions with a noise texture $T\in\{\textit{VG}, \textit{GN},$ $\textit{HD}, \textit{LN}, \textit{MN}\}$, we generate the image by using the texture with slightly different parameters for the character region ($p_c$) and the background region ($p_b$). This design introduces subtle differences between the character and background regions, thereby producing high-quality illusion images. Specifically, an illusion image $I_{\text{illusion}}$ with character region $\mathcal{D}_{\text{char}}$ is generated by
\begin{equation}
    I_{\text{illusion}}(x,y)=
    \begin{cases}
    T_{P_c}(x,y) & \text{if } (x,y)\in \mathcal{D}_\text{char} \\
    T_{P_b}(x,y) & \text{otherwise.}
    \end{cases}
\end{equation}

We further provide the description of each noise texture $T$:

\begin{itemize}
    \item \textbf{Vertical Gratings (VG)}: a texture composed of black-and-white vertical stripes. $T_{p_c}$ and $T_{p_b}$ differ in stripe width.
    \item \textbf{Gaussian Noise (GN)}: a texture using Gaussian noise. $T_{p_c}$ and $T_{p_b}$ differ in the base gray level used for Gaussian noise.
    \item \textbf{Halftone Dots (HD)}: a texture composed of randomly distributed dots. $T_{p_c}$ and $T_{p_b}$ differ in dot size.
    \item \textbf{Labyrinth Noise (LN)}: a maze-like binary texture generated by smoothing random noise and applying thresholding. $T_{p_c}$ and $T_{p_b}$ differ in the noise distribution.
    \item \textbf{Micro-text Noise (MN)}: a texture composed of randomly distributed micro symbols. $T_{p_c}$ uses `@' and `\&' symbols, while $T_{p_b}$ uses `\$', `\%', and `\#' symbols.
\end{itemize}

\begin{figure}[h]
  \centering
  \includegraphics[width=\linewidth]{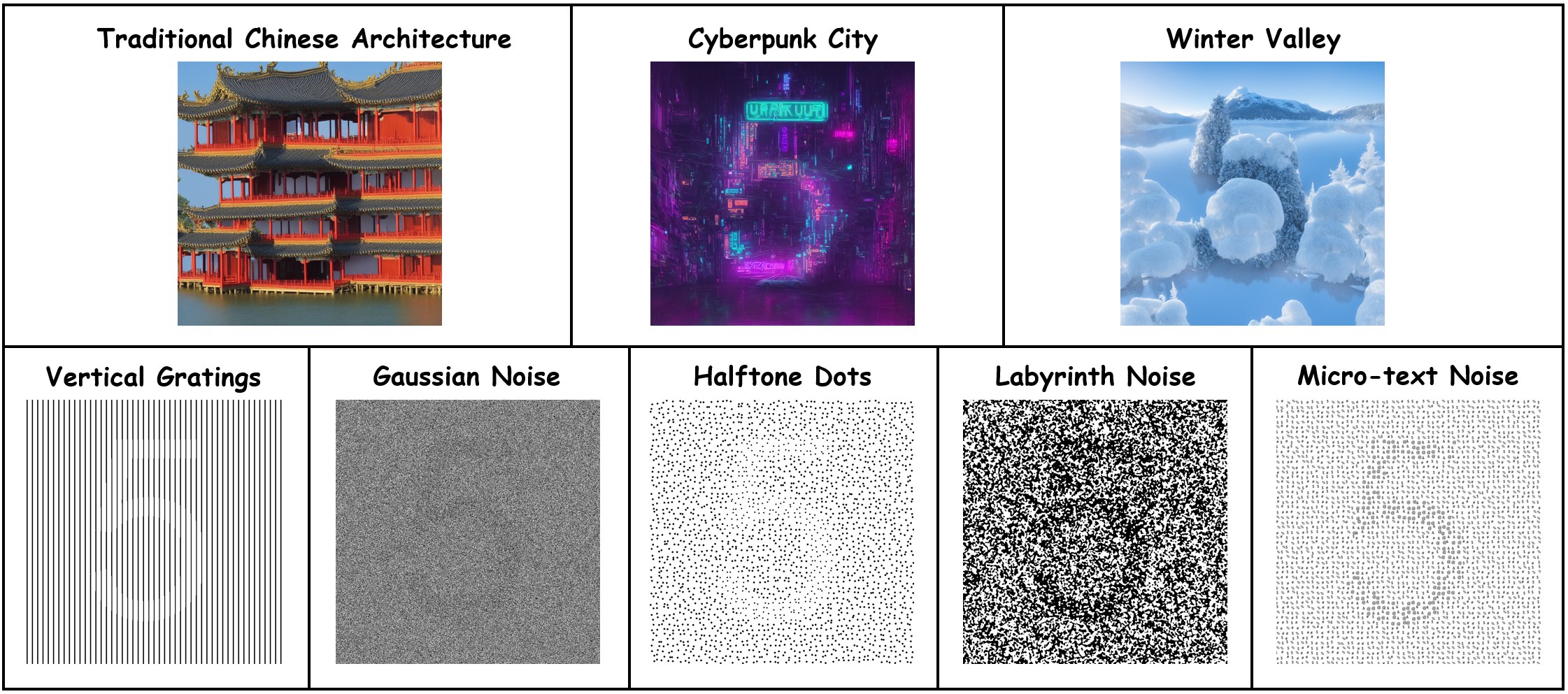}
  \caption{Examples of illusions with different backgrounds.}
  \label{fig:detail_dataset_example}
  \Description{Examples showing illusion images with different backgrounds.}
\end{figure}

In Figure \ref{fig:detail_dataset_example}, we provide illusion images with different backgrounds by embedding the digit `5' as an example.

\begin{figure}[h]
  \centering
  \includegraphics[width=0.8\linewidth]{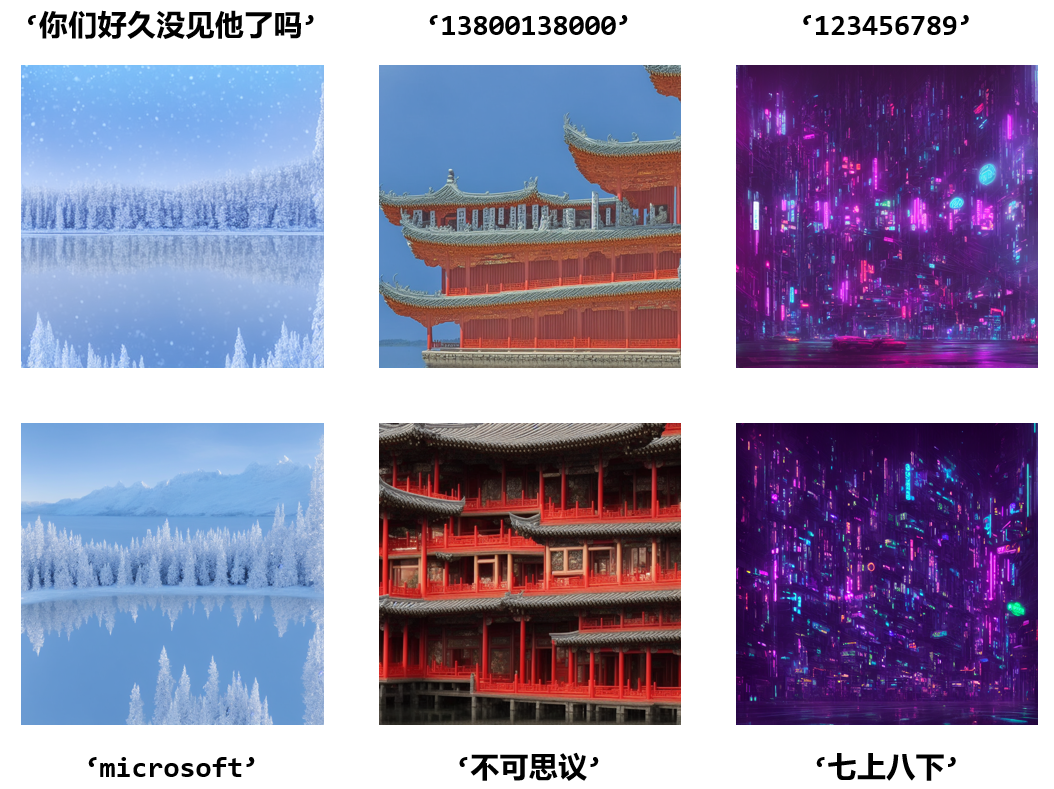}
  \caption{Examples of samples with unsatisfactory illusion quality.}
  \label{fig:fail_case}
  \Description{Examples of samples with unsatisfactory illusion quality.}
\end{figure}

After dataset construction, we further manually filtered out $5.5\%$ of samples with unsatisfactory illusion quality. All of these samples are semantic background images, where the character structures were completely destroyed during generation, making the hidden characters unrecognizable. We provide some examples in Figure \ref{fig:fail_case}. After removing this small portion of invalid samples, IlluChar retains only high-quality illusion images.

\subsection{Verification on the Effectiveness of Different Backgrounds}

We further verify the effectiveness of embedding characters into the proposed backgrounds. Specifically, we randomly select 100 hidden characters, and compare model performance on their original images and corresponding illusion images.

As shown in Table \ref{tab:backgrounds_effectiveness}, all types of backgrounds substantially degrade model performance, with accuracies dropping below $10\%$ in all cases. This result demonstrates the effectiveness of all proposed backgrounds in generating challenging illusion images.

\begin{table}
    \caption{Accuracies (\%) of six MLLMs on illusion images with different backgrounds.}
    \label{tab:backgrounds_effectiveness}
    \resizebox{\linewidth}{!}{
    \begin{tabular}{l | c ccc ccccc}
    \toprule
    \textbf{Models} & \textbf{Origin} & \multicolumn{3}{c}{\textbf{Semantic}} & \multicolumn{5}{c}{Noise} \\
    \cmidrule(lr){3-5} \cmidrule(lr){6-10}
    & & TC & CC & WV & VG & GN & HD & LN & MN \\
    \midrule
    Qwen3-VL-8B & 86.0 & 1.0 & 1.0 & 5.0 & 3.0 & 3.0 & 0.0 & 6.0 & 3.0 \\
    GLM-4.5V & 95.0 & 0.0 & 1.0 & 6.0 & 7.0 & 6.0 & 3.0 & 3.0 & 8.0 \\
    Qwen3-VL-235B-A22B & 96.0 & 0.0 & 1.0 & 8.0 & 4.0 & 6.0 & 1.0 & 4.0 & 4.0 \\
    GPT-5.2 & 81.0 & 2.0 & 1.0 & 9.0 & 6.0 & 2.0 & 1.0 & 2.0 & 8.0 \\
    Gemini-2.5-Pro & 93.0 & 2.0 & 4.0 & 7.0 & 5.0 & 5.0 & 5.0 & 8.0 & 9.0 \\
    Claude-Sonnet-4.5 & 85.0 & 0.0 & 0.0 & 9.0 & 0.0 & 2.0 & 1.0 & 0.0 & 1.0 \\
    \midrule
    \textbf{Human} & - & \textbf{99.0} & \textbf{100.0} & \textbf{100.0} & \textbf{100.0} & \textbf{98.0} & \textbf{99.0} & \textbf{98.0} & \textbf{100.0} \\
    \bottomrule
    \end{tabular}
    }
\end{table}

\section{Additional Attention Results on MLLMs}
\label{sec:appendixC}

In this section, we extend the attention analysis beyond CLIP-ViT: using attention maps from Qwen3-VL-8B-Instruct and black-box saliency maps (using RISE-style analysis method \cite{petsiuk2018rise}) from Gemini-2.5-Pro. 

\begin{figure}[h]
  \centering
  \includegraphics[width=\linewidth]{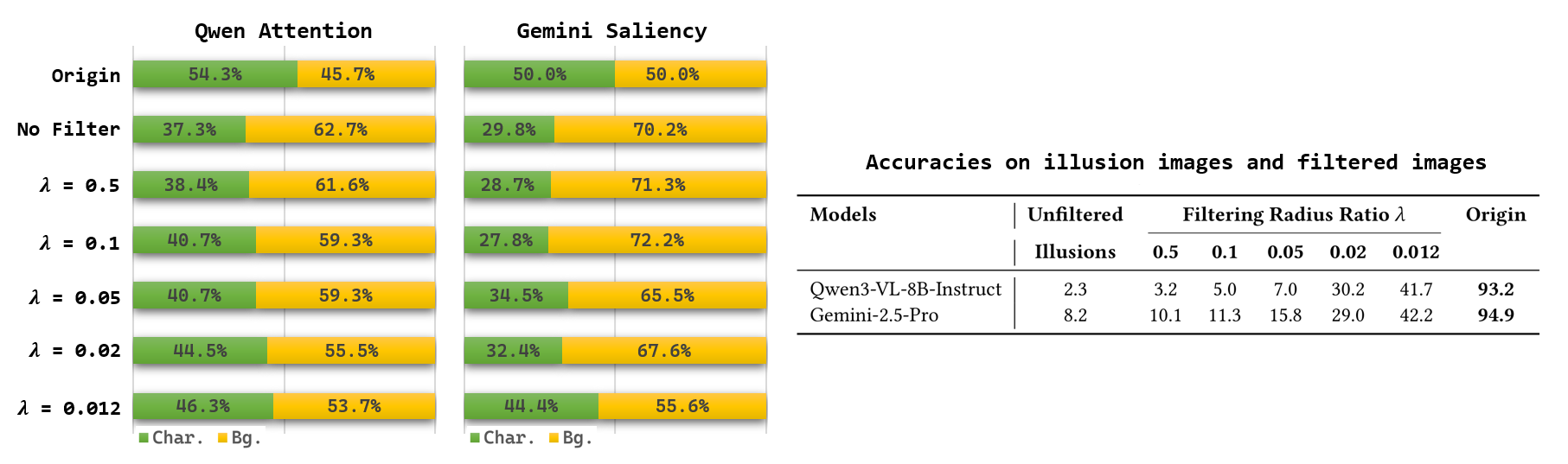}
  \caption{\textit{Left:} Attention distributions of two MLLMs. \textit{Right:} Recognition accuracies of two MLLMs.}
  \label{fig:attention_exp_mllms}
  \Description{Attention distribution and recognition accuracies of two models.}
\end{figure}

As shown in Figure \ref{fig:attention_exp_mllms}, both models exhibit the same trend: as the high-frequency signals in illusion images are suppressed, their attention shifts back toward the character regions and their recognition accuracy recovers. These results further show that the ``high-frequency attention bias'' is consistently observed in practical MLLMs.

\section{Human Experiments on Illusion Images}
\label{sec:appendixD}

In Section \ref{sec:dataset}, we recruited 10 participants to freely inspect and recognize images from IlluChar without any time limit, while observing their behaviors during recognition. We found that they achieved near-perfect accuracy, and all participants spontaneously adopted the two perception-adjustment strategies when facing illusion images, squinting and viewing the images from a distance.

Furthermore, we conducted an additional time-limited study with the same 10 participants, where each image was displayed for only 1 second, making it difficult for participants to adjust their perception. Under this setting, their accuracy dropped to $60.0\%$, validating that perceptual adjustment plays an important role in human recognition of visual illusions.

\section{Evaluation Details}
\label{sec:appendixE}

Our evaluation method combines strict string matching and GPT-based evaluation. 
Given a model response $y^*$ and the ground-truth answer $y$, we consider the following cases:

\begin{itemize}
    \item If $y \notin y^*$, the response does not contain the target character. We regard the response as incorrect.
    
    \item If $y \in y^*$, $|y| \ge 3$, and $y$ is not a common string that may naturally appear in model responses, we regard the response as correct.
    
    \item In all other cases, we employ GPT-4o as an evaluator to determine whether the response is correct.
\end{itemize}

To validate the reliability of this evaluation method, we randomly sample 1000 instances for manual annotation. The agreement between our method and human judgment reaches $99.5\%$, indicating that the method is reliable. We provide the prompt template for GPT-based evaluation here:

\begin{CJK}{UTF8}{gbsn}
\begin{Verbatim}[fontsize=\footnotesize,frame=single]
You are a strict evaluator. Your task is to determine whether the 
model's response correctly identifies the hidden number(s), letter
(s), word(s), or Chinese character(s) in the image.

You will be given a ground truth answer, which is the correct hidd-
en content, and a model response, which is the content identified 
by a specific model. You should compare the model response with the 
ground truth answer and decide if the model's identification is co-
rrect.
- [Correct]: If the model response exactly matches the ground truth 
answer.
- [Incorrect]: If the model response does not match the ground tru-
th answer.
Your output should only contain your evaluation result, either "Co-
rrect" or "Incorrect".

# Example 1
Ground Truth Answer: 5
Model Response: Looking at the image carefully, I can identify the 
hidden number: 5.
Evaluation: Correct
# Example 2
Ground Truth Answer: animal
Model Response: The hidden word in the image is "ANIMAL".
Evaluation: Correct
# Example 3
Ground Truth Answer: A
Model Response: The hidden letter in the image is B.
Evaluation: Incorrect
# Example 4
Ground Truth Answer: 我
Model Response: The hidden Chinese character in the image is 我.
Evaluation: Correct
# Example 5
Ground Truth Answer: 你好吗
Model Response: The hidden Chinese characters in the image are 我好
嘛.
Evaluation: Incorrect

Now it's your turn to evaluate.

Ground Truth Answer: [GROUND_TRUTH]
Model Response: [RESPONSE]
Evaluation:
\end{Verbatim}
\end{CJK}

\section{Prompts for the Experiments}
\label{sec:appendixF}

We provide prompt templates used in Section \ref{sec:experiments}. For Vanilla, Filtered, and Blur with Histogram methods, there is only one image input for the models. They utilize the same default prompt template:

\begin{Verbatim}[fontsize=\footnotesize,frame=single]
There is a [hidden_type] in the image, what is it ?
\end{Verbatim}

For Chain-of-Thought Prompting (CoT) method, we prompt the models to imagine visualizing the image using the proposed perceptual strategies--squinting eyes and viewing afar. The CoT prompt template is as follows:

\begin{Verbatim}[fontsize=\footnotesize,frame=single]
You are an expert in solving visual puzzles and optical illusions. 
Your task is to identify the hidden [hidden_type] embedded in the 
image. 

The image is designed as an optical illusion, where the character 
is subtly integrated into the semantic background or noise patter-
ns. 
To identify the hidden content, you can simulate human visual beha-
viors:
1. Imagine squinting your eyes or slightly blurring your vision. 
Ignore the sharp, high frequency details, textures and noise in the 
image.
2. Imagine viewing the image from a long distance. You can resize 
the image smaller in your mind to get a global view of the image.
You can combine the two strategies to enhance your perception of t-
he hidden character.

Now, please analyze the image carefully, and identify the hidden 
[hidden_type].
\end{Verbatim}

For SMSP, each sample is processed into four input images for the model. We provide the model with a brief description of the relationships among these images and prompt it to reason by jointly considering all four images. The prompt template is as follows:

\begin{Verbatim}[fontsize=\footnotesize,frame=single]
I provide four views of the SAME image, the original view and the 
global views. There is a SAME [hidden_type] embedded in these im-
ages, with the help of the views, what is it ?
\end{Verbatim}

\section{Image Processing Baselines}
\label{sec:appendixG}

In Section \ref{sec:experiments}, we compare SMSP with four baselines: Vanilla, CoT Prompting, Filtered, Blur with Histogram. Furthermore, we add additional comparison with 9 more image-processing baselines on Qwen3-VL-8B-Instruct.

\begin{table}[h]
    \caption{Comparing SMSP with image processing baselines.}
    \label{tab:baselines}
    \small
    \resizebox{\linewidth}{!}{
    \begin{tabular}{l | ccccc | c}
    \toprule
    \textbf{Baselines} & \textbf{Vanilla} & \textbf{Bilateral} & \textbf{Gaussian} & \textbf{Multi-} & \textbf{Multi-Resolution} & \textbf{SMSP} \\
     & & \textbf{Filtering} & \textbf{Filtering} & \textbf{Crop} & \textbf{Prompting} & \\
    \midrule
    Origin Avg. & 93.2 & 95.8 & 96.7 & 94.3 & 95.9 & \textbf{97.3} \\
    Noise Avg. & 16.7 & 38.5 & 59.4 & 40.6 & 59.0 & \textbf{87.3} \\
    Semantic Avg. & 3.8 & 25.2 & 24.7 & 16.1 & 32.2 & \textbf{77.2} \\
    \midrule
    \textbf{Baselines} & \textbf{Median} & \textbf{Image} & \textbf{OCR-Style} & \textbf{Learned} & \textbf{Text-Based} & \textbf{SMSP} \\
     & \textbf{Denoising} & \textbf{Pyramid} & \textbf{Preprocessing} & \textbf{Preprocessor} & \textbf{Perceptual Prompts} & \\
    \midrule
    Origin Avg. & 89.7 & 96.9 & 90.0 & 97.0 & 93.2 & \textbf{97.3} \\
    Noise Avg. & 32.0 & 61.1 & 9.5 & 15.1 & 17.5 & \textbf{87.3} \\
    Semantic Avg. & 31.0 & 30.2 & 10.1 & 11.7 & 3.9 & \textbf{77.2} \\
    \bottomrule
    \end{tabular}
    }
\end{table}

As shown in Table \ref{tab:baselines}, SMSP consistently outperforms these baselines on illusion images with both backgrounds, further indicating its effectiveness.

\section{Additional Results on MLLMs of Different Sizes}

In Section \ref{sec:experiments}, we evaluate SMSP on MLLMs from different families. The results demonstrate its transferability across diverse model architectures. Furthermore, we investigate SMSP's scalability by testing models of varying sizes from the Qwen3-VL series \cite{Qwen3-VL}. Specifically, our evaluation includes models with 2B, 4B, 8B, 30B-A3B, and 235B-A22B parameters, covering a wide range of model scales.

As shown in Table \ref{tab:main_results2}, SMSP consistently outperforms baseline methods across all evaluated models, substantially improving their performance on all types of illusion images, which indicates the robustness of SMSP across models of different scales.

\begin{table}[t]
    \caption{Accuracies (\%) of Qwen-series models with different methods on IlluChar.}
    \label{tab:main_results2}
    \small
    \resizebox{\linewidth}{!}{
    \begin{tabular}{l | cccc cccc cccc}
    \toprule
    \textbf{Method} 
    & \multicolumn{4}{c}{\textbf{Origin}} 
    & \multicolumn{4}{c}{\textbf{Noise}}
    & \multicolumn{4}{c}{\textbf{Semantic}} \\
    \cmidrule(lr){2-5} \cmidrule(lr){6-9} \cmidrule(lr){10-13}
    & L & M & S & Avg. & L & M & S & Avg. & L & M & S & Avg. \\
    \midrule
    & \multicolumn{12}{c}{\colorbox{yellow!10}{\textbf{Qwen3-VL-2B-Instruct}}} \\
    \cmidrule(lr){2-13}
    Vanilla & 67.2 & 97.9 & 96.4 & 83.6 & 0.5 & 14.6 & 48.4 & 17.5 & 1.3 & 8.9 & 2.7 & 3.4 \\
    CoT & 67.2 & 95.0 & \textbf{98.6} & 83.4 & 0.7 & 16.1 & 44.3 & 16.8 & 1.0 & 11.3 & 4.0 & 4.0 \\
    Filtered & 84.5 & \textbf{99.3} & 83.6 & 88.3 & 6.5 & 23.7 & 9.7 & 12.1 & 25.9 & 78.7 & \textbf{80.0} & 43.9 \\
    Blur with Histogram & 80.2 & 88.6 & 25.0 & 67.4 & 33.6 & 55.9 & 2.3 & 31.1 & 32.9 & 46.1 & 14.7 & 35.1 \\
    \textbf{SMSP (Ours)} & \textbf{88.8} & 98.6 & 97.1 & \textbf{93.8} & \textbf{64.8} & \textbf{92.3} & \textbf{86.3} & \textbf{78.2} & \textbf{61.8} & \textbf{90.8} & 70.7 & \textbf{70.2} \\
    \midrule
    
    & \multicolumn{12}{c}{\colorbox{purple!10}{\textbf{Qwen3-VL-4B-Instruct}}} \\
    \cmidrule(lr){2-13}
    Vanilla  & 75.0 & 99.3 & \textbf{99.3} & 88.3 & 1.0 & 8.7 & 45.7 & 15.4 & 2.3 & 7.4 & 8.0 & 4.1 \\
    CoT      & 74.1 & \textbf{100.0} & 98.6 & 87.9 & 1.5 & 8.0 & 43.7 & 14.8 & 2.4 & 7.8 & 6.7 & 4.2 \\
    Filtered & 88.8 & \textbf{100.0} & 85.0 & 90.8 & 12.2 & 28.9 & 12.6 & 16.9 & 28.9 & 75.9 & \textbf{80.0} & 45.1 \\
    Blur with Histogram & 89.2 & 95.0 & 23.6 & 72.9 & 39.5 & 61.6 & 1.9 & 35.2 & 43.1 & 48.6 & 13.3 & 42.5 \\
    \textbf{SMSP (Ours)}
             & \textbf{96.6} & 99.3 & 97.9 & \textbf{97.7} & \textbf{70.9} & \textbf{92.6} & \textbf{86.9} & \textbf{81.2} & \textbf{65.1} & \textbf{77.0} & 78.7 & \textbf{69.2} \\
    \midrule
    
    & \multicolumn{12}{c}{\colorbox{cyan!5}{\textbf{Qwen3-VL-8B-Instruct}}} \\
    \cmidrule(lr){2-13}
    Vanilla  & 87.1 & 99.3 & 97.1 & 93.2 & 0.6 & 11.4 & 48.7 & 16.7 & 2.4 &  6.4 & 6.7 & 3.8 \\
    CoT      & 86.6 & 99.3 & 97.9 & 93.2 & 1.0 & 12.3 & 49.9 & 17.5 & 1.6 & 8.9 & 6.7 & 3.9 \\
    Filtered & 94.4 & 99.3 & 97.9 & 96.7 & 20.9 & 59.1 & 17.1 & 30.3 & 26.0 & 81.9 & 80.0 & 44.8 \\
    Blur with Histogram & 90.9 & 95.0 & 10.0 & 69.9 & 42.6 & 66.0 & 1.4 & 37.7 & 41.8 & 42.9 & 6.7 & 39.6 \\
    \textbf{SMSP (Ours)}
             & \textbf{95.3} & \textbf{100.0} & \textbf{97.9} & \textbf{97.3} & \textbf{79.1} & \textbf{95.4} & \textbf{93.0} & \textbf{87.3} & \textbf{74.0} & \textbf{83.7} & \textbf{82.7} & \textbf{77.2} \\
    \midrule
    
    & \multicolumn{12}{c}{\colorbox{orange!10}{\textbf{Qwen3-VL-30B-A3B-Instruct}}} \\
    \cmidrule(lr){2-13}
    Vanilla  & 93.1 & 99.3 & 97.9 & 96.1 & 1.4 & 34.1 & 62.4 & 27.0 & 2.0 & 3.2 & 2.7 & 2.4 \\
    CoT      & 92.2 & 99.3 & 97.1 & 95.5 & 1.8 & 31.3 & 61.1 & 26.1 & 2.7 & 5.7 & 4.0 & 3.6 \\
    Filtered & 93.5 & 100.0 & 90.0 & 94.3 & 22.5 & 40.6 & 17.6 & 26.1 & 21.0 & 85.1 & 88.0 & 42.9 \\
    Blur with Histogram & 92.2 & 95.7 & 37.1 & 78.1 & 37.4 & 61.6 & 2.0 & 34.3 & 43.1 & 56.4 & 13.3 & 44.5 \\
    \textbf{SMSP (Ours)}
             & \textbf{95.7} & \textbf{100.0} & \textbf{100.0} & \textbf{98.0} & \textbf{71.3} & \textbf{95.3} & \textbf{89.6} & \textbf{82.9} & \textbf{64.5} & \textbf{92.2} & \textbf{89.3} & \textbf{73.7} \\
    \midrule

    & \multicolumn{12}{c}{\colorbox{blue!10}{\textbf{Qwen3-VL-235B-A22B-Instruct}}} \\
    \cmidrule(lr){2-13}
    Vanilla  & 95.3 & 100.0 & 99.3 & 97.7 & 0.9 & 16.4 & 55.4 & 20.0 & 2.7 & 7.8 & 6.7 & 4.4 \\
    CoT      & 94.8 & 100.0 & 100.0 & 97.7 & 1.7 & 15.9 & 52.9 & 19.6 & 3.0 & 6.4 & 9.3 & 4.4 \\
    Filtered & 97.4 & 100.0 & 96.4 & 97.9 & 24.4 & 73.1 & 28.6 & 38.9 & 33.9 & 88.3 & 93.3 & 52.7 \\
    Blur with Histogram & 94.8 & 99.3 & 21.4 & 76.0 & 42.9 & 80.0 & 3.1 & 42.2 & 59.5 & 60.6 & 17.3 & 56.8 \\
    \textbf{SMSP (Ours)}
             & \textbf{99.6} & \textbf{100.0} & \textbf{100.0} & \textbf{99.8} & \textbf{79.7} & \textbf{98.0} & \textbf{94.0} & \textbf{88.6} & \textbf{78.2} & \textbf{89.7} & \textbf{94.7} & \textbf{82.4} \\

    \bottomrule
    \end{tabular}
    }
\end{table}

\section{Boundary Parameter Selection}
\label{sec:appendixI}

\begin{figure}
  \centering
  \includegraphics[width=0.9\linewidth]{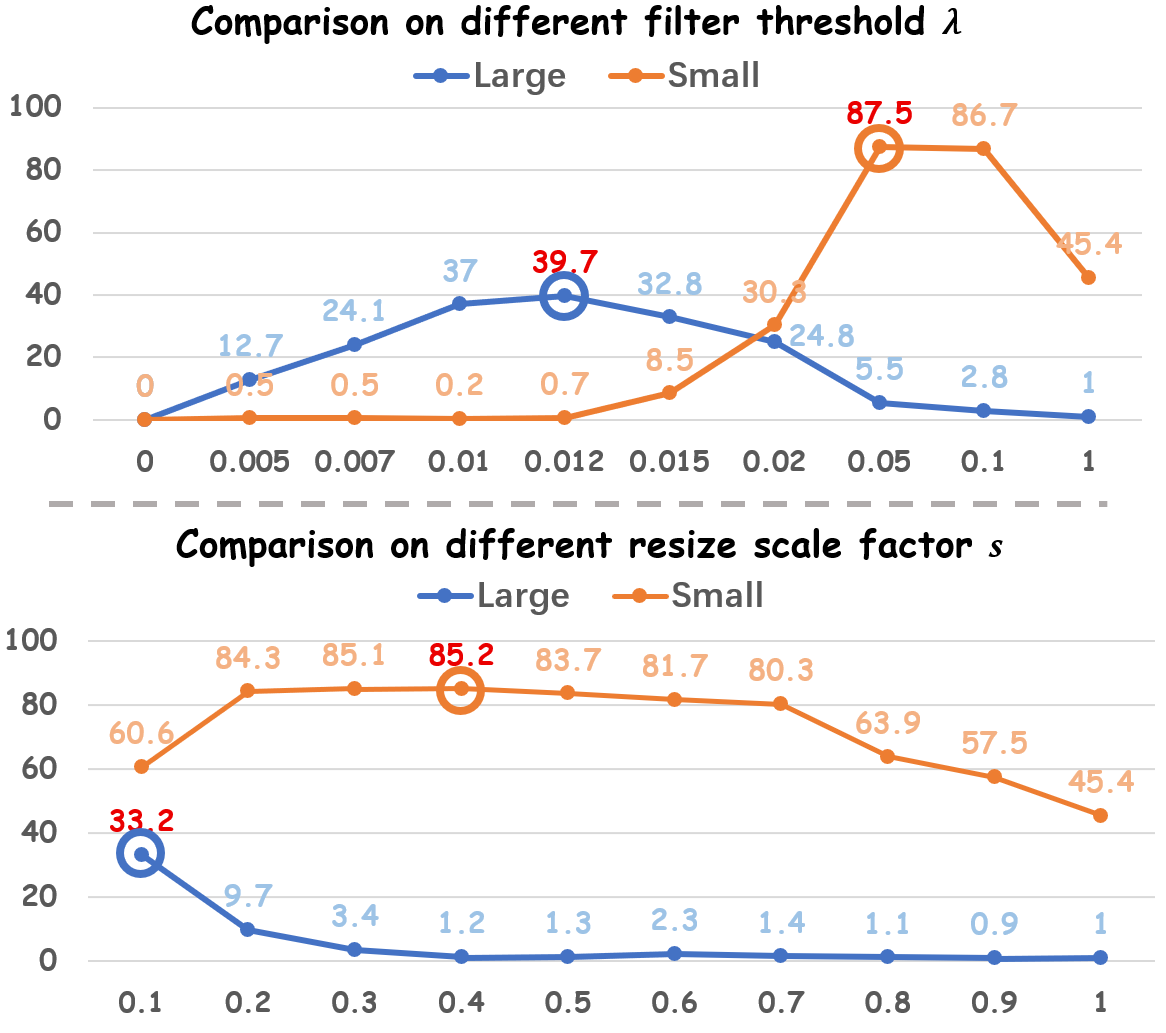}
  \caption{Empirical selection of SMSP boundary parameters. Accuracies are evaluated on samples with the largest and smallest hidden characters.}
  \label{fig:param_sel}
  \Description{Two plots showing model accuracy under different parameter settings.}
\end{figure}

To determine the optimal boundary parameters $(\lambda_1, s_1)$ and $(\lambda_K, s_K)$, we utilize a validation set containing hidden characters with extreme scales (i.e., the largest and smallest characters). This allows us to identify the most effective processing strengths for the strongest and weakest variants, respectively. We conduct a grid search on the validation set to find the best parameters. As shown in Figure \ref{fig:param_sel}, $\lambda_1=0.012$ and $s_1=0.1$ yield the best performance on samples with the largest hidden characters, while $\lambda_K=0.05$ and $s_K=0.4$ perform best on the smallest character samples.

We further examine whether the selected boundary parameters remain effective across different models and illusion settings. As shown in Table \ref{tab:boundary_selection}, we perturb each boundary parameter on 2 OOD illusion datasets across 3 models, and the selected settings consistently achieve strong performance. This indicates that the boundary parameters are robust and can be applied without model-specific re-tuning. Therefore, we use this setting for all experiments.

\begin{table}
    \caption{Validation of the chosen boundary parameters.}
    \label{tab:boundary_selection}
    \small
    \resizebox{\linewidth}{!}{
    \begin{tabular}{c | l | cccc cccc ccc cccc}
    \toprule
    \textbf{Models} & \textbf{Datasets} 
    & \multicolumn{4}{c}{\textbf{$\lambda_1$}} 
    & \multicolumn{4}{c}{\textbf{$\lambda_K$}}
    & \multicolumn{4}{c}{\textbf{$s_1$}}
    & \multicolumn{3}{c}{\textbf{$s_K$}} \\
    \cmidrule(lr){3-6} \cmidrule(lr){7-10} \cmidrule(lr){11-13}
    \cmidrule(lr){14-17}
    & & 0.01 & \boxed{0.012} & 0.015 & 0.02 & 0.02 & \boxed{0.05} & 0.1 & 1 & \boxed{0.1} & 0.2 & 0.3 & 0.2 & 0.3 & \boxed{0.4} & 0.5 \\
    \midrule
    Qwen3- & IllusoryVQA & \textbf{79.7} & \textit{\underline{79.0}} & \textit{\underline{79.0}} & 77.7 & \textit{\underline{78.5}} & \textbf{79.0} & 77.5 & 66.5 & \textbf{79.0} & \textit{\underline{78.3}} & 77.5 & \textit{\underline{78.7}} & 77.8 & \textbf{79.0} & 78.0\\
    \cmidrule(lr){2-17}
    VL-8B & HatefulIllusion & 64.0 & \textbf{71.0} & \textit{\underline{67.0}} & 66.3 & \textit{\underline{63.0}} & \textbf{71.0} & 60.0 & 41.8 & \textbf{71.0} & 64.0 & \textit{\underline{68.0}} & \textit{\underline{66.8}} & 64.5 & \textbf{71.0} & 65.8 \\
    \midrule
    GPT-5.2 & IllusoryVQA & 68.0 & \textbf{69.1} & 67.7 & \textit{\underline{68.8}} & \textbf{69.2} & \textit{\underline{69.1}} & 66.3 & 57.1 & \textbf{69.1} & \textit{\underline{67.7}} & \textit{\underline{67.7}} & 65.1 & \textit{\underline{68.6}} & \textbf{69.1} & \textbf{69.1} \\
    \cmidrule(lr){2-17}
    & HatefulIllusion & 24.3 & \textbf{27.0} & \textit{\underline{24.5}} & 24.3 & \textbf{27.3} & \textit{\underline{27.0}} & 21.5 & 16.0 & \textbf{27.0} & 26.8 & 25.3 & 26.1 & \textit{\underline{26.3}} & \textbf{27.0} & 26.1 \\
    \midrule
    Gemini- & IllusoryVQA & \textit{\underline{78.2}} & \textbf{80.3} & 77.3 & 77.6 & 77.0 & \textbf{80.3} & \textit{\underline{79.6}} & 73.8 & \textbf{80.3} & \textit{\underline{80.1}} & 79.5 & 77.4 & 79.0 & \textbf{80.3} & \textit{\underline{79.5}}\\
    \cmidrule(lr){2-17}
    2.5-Pro & HatefulIllusion & 62.8 & \textbf{70.2} & \textit{\underline{67.5}} & 63.5 & \textbf{70.6} & \textit{\underline{70.2}} & 60.7 & 53.8 & \textbf{70.2} & \textit{\underline{66.3}} & 65.3 & 67.5 & 66.0 & \textbf{70.2} & \textit{\underline{67.8}} \\
    \bottomrule
    \end{tabular}
    }
\end{table}

\section{Case Study}

\begin{figure}
  \centering
  \includegraphics[width=\linewidth]{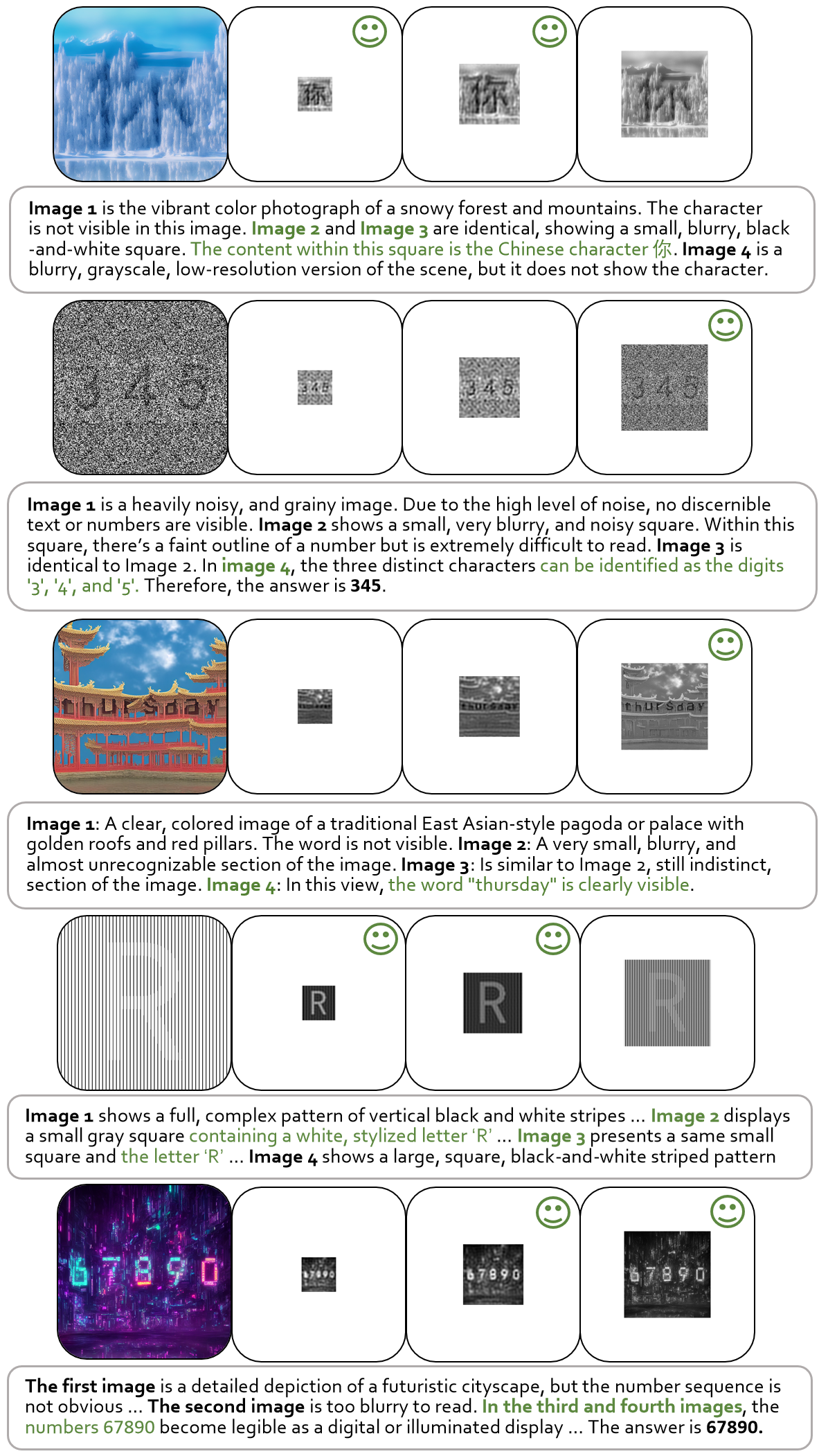}
  \caption{Examples of SMSP on illusions with more diverse settings.}
  \label{fig:case_appendix}
  \Description{More examples showing SMSP's effectiveness.}
\end{figure}

Apart from the two cases presented in Section \ref{sec:case_study}, we provide additional examples here to further illustrate the effectiveness of SMSP. As shown in Figure \ref{fig:case_appendix}, across various types of illusion images--including variations in hidden character type, character size, and background type--SMSP consistently generates clear perception-adjusted variants. The model is then able to identify more informative images and successfully recognize the hidden content.

\begin{figure}[t]
  \centering
  \includegraphics[width=0.93\linewidth]{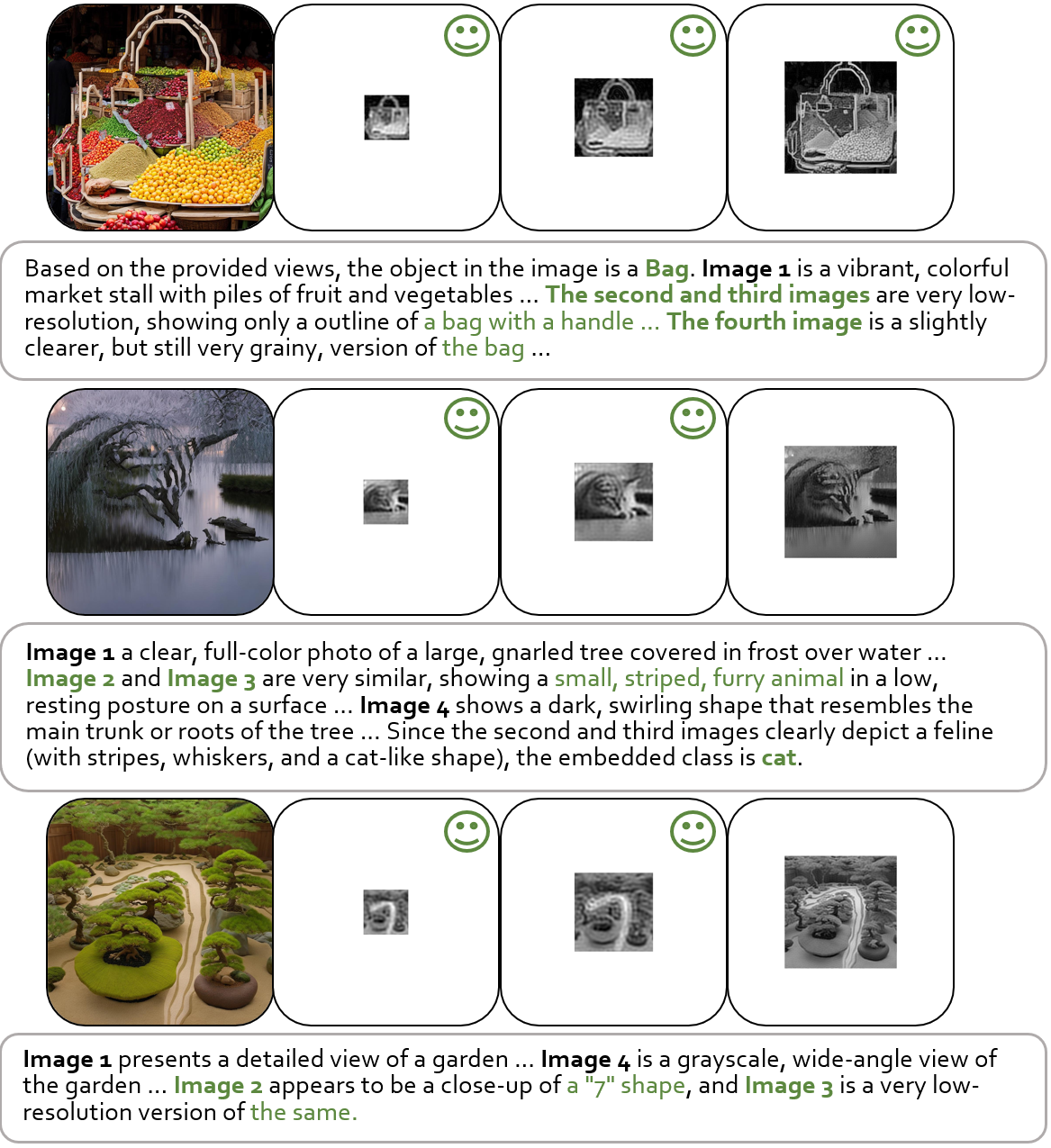}
  \caption{Examples of SMSP on samples in IllusoryVQA \cite{rostamkhani2025illusory}.}
  \label{fig:case_appendix2}
  \Description{More examples showing SMSP's effectiveness on more diverse hidden patterns.}
\end{figure}

In Figure \ref{fig:case_appendix2}, we further present examples of illusions with more diverse hidden patterns selected from the IllusoryVQA \cite{rostamkhani2025illusory} dataset. The results further demonstrate the effectiveness of SMSP and highlight its generalizability across a variety of illusion images.

\section{Limitations in Fine-tuning Strategy}

We further explore model fine-tuning as a potential mitigation strategy. Following \cite{rostamkhani2025illusory}, we fine-tune Qwen3-VL-4B-Instruct using similar hyperparameters (the parameters are provided in Table \ref{tab:SFT_params}). We use 2.5k illusion images from IlluChar for training and reserve the remaining data (approximately 1.1k) as the test set. We evaluate the fine-tuned model on both the IlluChar test set and illusion images with other hidden patterns. We then compare it with the vanilla 4B model and the vanilla model enhanced with SMSP.

As shown in Table \ref{tab:SFT}, the model fine-tuned on IlluChar indeed achieves improved performance on IlluChar itself, but still lags behind SMSP. Furthermore, on illusion images with other hidden patterns, the fine-tuned model performs similarly to the vanilla model, suggesting its limited generalization. These results indicate that perception-based improvements cannot be reliably obtained through additional model training alone, highlighting the necessity of explicitly incorporating perceptual strategies.

\begin{table}[h]
    \caption{Hyperparameters used for fine-tuning.}
    \label{tab:SFT_params}
    \resizebox{0.5\linewidth}{!}{
    \begin{tabular}{c | c | c}
    \toprule
    \multicolumn{2}{c|}{\textbf{Hyperparameter}} & \textbf{Value} \\
    \midrule
    \multicolumn{2}{c|}{\textbf{Learning Rate}} & $1e-5$ \\
    \multicolumn{2}{c|}{\textbf{Batch Size (per device)}} & $8$ \\
    \multicolumn{2}{c|}{\textbf{Epochs}} & $2$ \\
    \multicolumn{2}{c|}{\textbf{Optimizer}} & AdamW \\
    \midrule
    \textbf{Lora \cite{hu2022lora}} & \textbf{r} & $64$ \\
    & \textbf{lora\_alpha} & $16$ \\
    & \textbf{target\_modules} & all-linear \\
    & \textbf{dropout} & 0 \\
    \bottomrule
    \end{tabular}
    }
\end{table}

\section{Broader Impact}

Our work introduces a novel perception-inspired paradigm for improving MLLMs' understanding of images by simulating human perceptual strategies. Compared with methods that rely on retraining the original model, our method is simple and training-free, requiring no additional training data and enabling plug-and-play deployment.

Beyond visual illusions, this paradigm may also inspire new perception-driven strategies for more complex tasks, including fine-grained image analysis and video understanding, which enables its broader applications across diverse visual understanding tasks.

\begin{table}
    \caption{Accuracies (\%) of the vanilla 4B model, the fine-tuned model, and the SMSP method.}
    \label{tab:SFT}
    \resizebox{0.85\linewidth}{!}{
    \begin{tabular}{l | c | cccc}
    \toprule
    \textbf{Method} & \textbf{IlluChar} & \textbf{Animals} & \textbf{Fashion-} & \textbf{MNIST} & \textbf{Harmful} \\
    & & & \textbf{MNIST} & & \textbf{Pattern} \\
    \midrule
    Vanilla & 11.6 & 33.5 & 8.5 & 17.5 & 1.0 \\
    Fine-tuning & 22.9 & 31.5 & 12.0 & 19.0 & 1.5 \\
    \textbf{SMSP} & \textbf{77.9} & \textbf{92.5} & \textbf{48.0} & \textbf{90.5} & \textbf{25.0} \\
    \bottomrule
    \end{tabular}
    }
\end{table}

\section{Limitations and Future Work}

Although SMSP significantly improves model performance on illusion images across all evaluated settings, it relies on predefined perceptual parameters and lacks the ability to dynamically adapt to different input images. In addition, the use of fixed parameters increases the number of input images by a factor of four. Although Section \ref{sec:ablation} shows that the resulting time overhead is minimal, it still incurs additional token costs. Future work could investigate training lightweight networks to predict optimal perceptual parameters or dynamically select the most suitable perceptually adjusted variant for each input image.

In addition, as discussed, our work only explores aligning the model's perceptual strategy with human perception in the context of visual illusions. Future work could build upon the perceptual paradigm introduced by SMSP to design additional visual perception strategies for MLLMs, enabling them to better handle a broader range of tasks.


\end{document}